%% file: paper.tex
\documentclass[]{TEAI}
\usepackage{helvet}

\usepackage{amsmath} 
\usepackage{natbib}
\usepackage{graphicx}
\usepackage{subcaption} 

\usepackage[toc,page,header]{appendix}
\usepackage[utf8]{inputenc} 
\usepackage[T1]{fontenc}    
\usepackage{hyperref}       
\usepackage{url}            
\usepackage{booktabs}       
\usepackage{lmodern}        
\usepackage{amsfonts}       
\usepackage{nicefrac}       
\usepackage{microtype}      
\usepackage{wrapfig}

\usepackage{amssymb}  
\usepackage{fontawesome}  
\usepackage{url}  

\usepackage{titletoc}

\usepackage{tikz}  
\usepackage{comment}  
\usepackage{tabularx}  
\usepackage{booktabs}  

\usepackage{minitoc}

\usepackage{booktabs}
\usepackage{array}
\usepackage{etoolbox}

\definecolor{lightblue}{RGB}{200, 230, 255}  
\definecolor{headerblue}{RGB}{150, 200, 255} 

\usepackage{pgfplots}
\usepackage[utf8]{inputenc} 
\usepackage[T1]{fontenc}    
\usepackage{hyperref}       
\usepackage{url}            
\usepackage{booktabs}       
\usepackage{amsfonts}       
\usepackage{nicefrac}       
\usepackage{microtype}      
\usepackage{xcolor}         
\usepackage{graphicx}
\usepackage{float}
\usepackage{comment}
\usepackage{multirow} 
\usepackage{amsmath} 
\usepackage{makecell} 
\usepackage{siunitx}  
\usepackage{tikz}
\usepackage{pgf-pie} 
\usepackage{subcaption}
\usepackage{wrapfig}
\usepackage[export]{adjustbox}

\usepackage{ragged2e}      
\usepackage{tabularx}       
\usepackage{array}          
\usepackage{caption}        
\usepackage{enumitem}
\usepackage{pifont}
\usepackage[hang,flushmargin]{footmisc} 

\usepackage{tcolorbox}

\usepackage{tcolorbox}    
\tcbuselibrary{breakable}  
\tcbuselibrary{skins}      

\usepackage{tabularx}
\usepackage{listings}

\title{{\fontsize{16pt}{22pt}\selectfont \textsc{CT-1}: Vision-Language-Camera Models Transfer Spatial Reasoning Knowledge to Camera-Controllable Video Generation}}

\author{
    Haoyu Zhao\textsuperscript{1,$\spadesuit$},
    Zihao Zhang\textsuperscript{1},
    Jiaxi Gu\textsuperscript{2,$\diamondsuit$},
    Haoran Chen\textsuperscript{1},
    Qingping Zheng\textsuperscript{3}, \\
    Pin Tang\textsuperscript{4},
    Yeyin Jin\textsuperscript{2},
    Yuang Zhang\textsuperscript{2},
    Junqi Cheng\textsuperscript{2},  
    Zenghui Lu\textsuperscript{2}, \\
    Peng Shu\textsuperscript{2},
    Zuxuan Wu\textsuperscript{1,$\dagger$},
    Yu-Gang Jiang\textsuperscript{1,$\dagger$}
}

\affiliation[1]{\mbox{Fudan University}}
\affiliation[2]{\mbox{Tencent}}
\affiliation[3]{\mbox{Xiamen University}}
\affiliation[4]{\mbox{Shanghai Jiao Tong University}}

\abstract{
\begin{abstract}

Camera-controllable video generation aims to synthesize videos with flexible and physically plausible camera movements. However, existing methods either provide imprecise camera control from text prompts or rely on labor-intensive manual camera trajectory parameters, limiting their use in automated scenarios.
To address these issues, we propose a novel Vision-Language-Camera model, termed CT-1 (\textsc{Camera Transformer 1}), a specialized model designed to transfer spatial reasoning knowledge to video generation by accurately estimating camera trajectories. Built upon vision-language modules and a Diffusion Transformer model, CT-1 employs a Wavelet-based Regularization Loss in the frequency domain to effectively learn complex camera trajectory distributions.
These trajectories are integrated into a video diffusion model to enable spatially aware camera control that aligns with user intentions.
To facilitate the training of CT-1, we design a dedicated data curation pipeline and construct CT-200K, a large-scale dataset containing over 47M frames. 
Experimental results demonstrate that our framework successfully bridges the gap between spatial reasoning and video synthesis, yielding faithful and high-quality camera-controllable videos and improving camera control accuracy by 25.7\% over prior methods.
\end{abstract}
}

\correspondence{Zuxuan Wu at \email{zxwu@fudan.edu.cn}, and Yu-Gang Jiang at \email{ygj@fudan.edu.cn}}
\checkdata[Code]{\href{https://github.com/gulucaptain/Camera-Transformer-1}{https://github.com/gulucaptain/Camera-Transformer-1}}
\checkdata[Website]{\href{https://gulucaptain.github.io/Camera-Transformer-1/}{https://gulucaptain.github.io/Camera-Transformer-1/}}

\begin{document}
\maketitle
\renewcommand{\thefootnote}{}
\footnotetext{$^\spadesuit$This work was completed during a research internship at Tencent as part of the Qingyun Program.\\$^\diamondsuit$Project leader.\\$^\dagger$Corresponding authors.}
\renewcommand{\thefootnote}{\arabic{footnote}}


\vspace{-1.5em}

\input{section/introduction}

\input{section/related}

\input{section/method}

\input{section/results}
\input{section/conclusion}

\clearpage

\bibliographystyle{plainnat}
\bibliography{main}

\clearpage

\newpage



\input{section/appendix}

\end{document}

%% file: section/introduction.tex
\section{Introduction}

\begin{figure*}[t!]
    \centering
    \subfloat[We propose CT-1, a Vision–Language–Camera (VLC) model that takes a scene image and a textual description encoding user intent as inputs to infer dynamic camera motion trajectories. By integrating CT-1 with a controllable video generation model, our framework enables the transfer of spatial reasoning knowledge from the VLC model to the generative models.]
    {\includegraphics[width=1.0\linewidth]{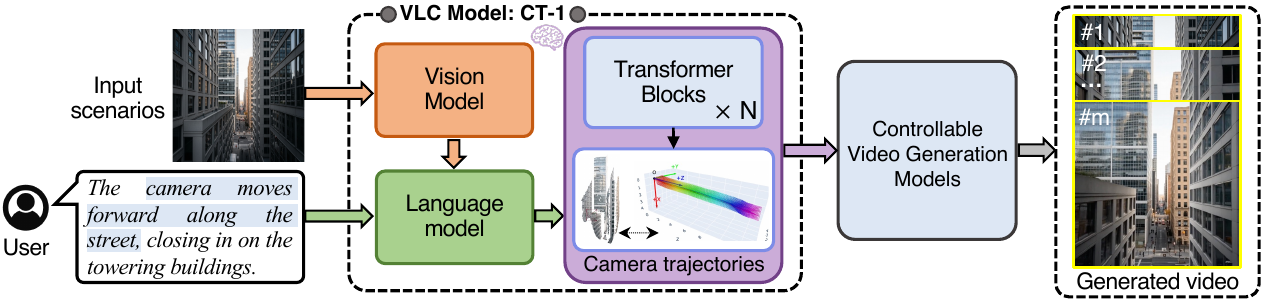}}\\
    \subfloat[We evaluate our proposed framework on out-of-distribution (OOD) scenarios and compare it with existing state-of-the-art video generation models, including Wan2.2~\cite{wan2025wan} and the camera-controlled model Gen3C~\cite{ren2025gen3c}.]{
    \hspace{-5pt}
    \includegraphics[width=1.0\linewidth]{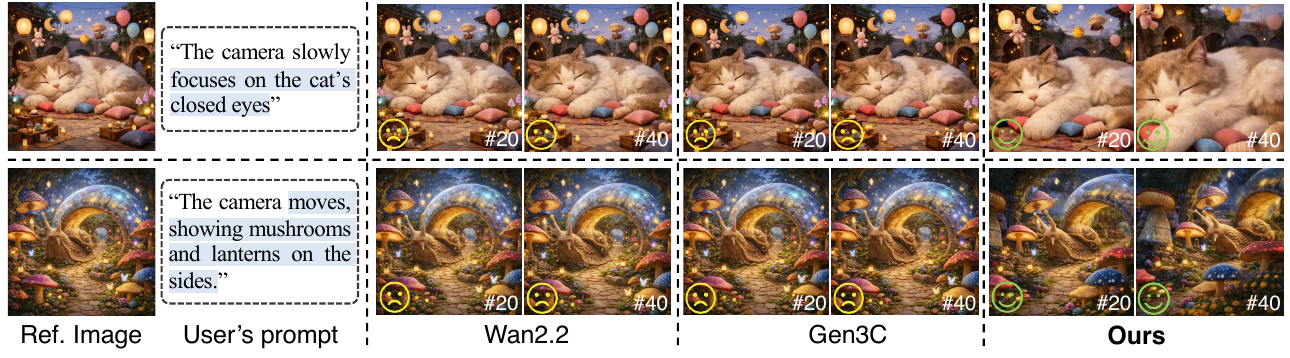}}%
    \caption{A high-level overview of CT-1’s architecture, its integration with the video generation model, and comparisons.}
    \label{fig:intro}
    \vspace{-10pt}
\end{figure*}

Camera-controllable video generation has emerged as a central research topic, leading to rapid advances in diffusion-based video generation. Camera motion is essential for enhancing the narrative quality and dynamic realism of videos, with broad applications in commercial video creation, animation, and world models. Based on the form of control signals, existing approaches can be divided into high-level semantic control~\cite{wan2025wan,li2025training} and explicit parameterized control~\cite{wang2024motionctrl,he2025cameractrl,he2025cameractrlII}.
The former typically relies on text prompts to specify the desired camera motion in a holistic manner, whereas the latter employs camera trajectory parameters as conditioning signals, thereby enabling more precise control over camera motion.

Despite substantial advances in camera control, generating camera trajectories that are consistent with both user intention and scene context remains a fundamental challenge. For explicit parameterized control methods, manually specifying camera trajectory parameters that match the desired intent and scenario is highly labor-intensive, which limits their scalability in large-scale automated applications. By contrast, current high-level semantic control methods, which directly employ video foundation models, often suffer from imprecise control and may even fail to follow camera motion instructions, as shown by the Wan2.2~\cite{wan2025wan} examples in Fig.~\ref{fig:intro} (b). Consequently, existing methods still fall short of fundamentally solving camera-controllable video generation across diverse user intentions and scene scenarios.

To this end, we explore a novel model that automatically generates camera trajectories by leveraging scene-level spatial knowledge and user intention. We term such models Vision-Language-Camera (VLC) models. 
A key advantage of VLC models is their ability to \textit{jointly reason about visual content and semantic cues to predict future spatiotemporal camera poses}, thereby improving semantic alignment between camera trajectories and the generated video content, as shown in Fig.~\ref{fig:intro} (a).
However, building such models remains challenging. The two main difficulties lie in training data construction and model design.
First, data collection and curation are the ``unsung heroes'' of vision–language models, including the VLC.
Second, unlike prior VLM methods for coordinate or bounding-box prediction~\cite{du2022learning,chen2023large}, which estimate spatial locations from explicit visual semantics, VLC requires predicting a temporally coherent camera trajectory over time. This is particularly challenging because the reference image provides only indirect cues about plausible camera motion, rather than explicit spatial targets or supervision.

In this paper, we introduce a novel framework to generate camera-controllable videos by bridging high-level user intention and low-level scenario pixels through a dedicated Vision-Language-Camera model, termed CT-1 (Camera Transformer 1). Rather than relying on manually specified camera trajectories or vague text-only guidance, CT-1 learns to infer scene-aware and intention-aligned camera trajectories, thereby transferring spatial reasoning ability into diffusion-based video generation.
Specifically, CT-1 is built upon a vision-language model that encodes camera-aware semantic cues into camera-context tokens, which are further integrated into diffusion transformer blocks to model a distribution over plausible camera trajectories. 
To encourage temporally smooth and physically stable camera motion, we further introduce a Wavelet-based Regularization Loss in the frequency domain. 
Based on the predicted trajectories, our framework enables robust controllable video generation across diverse scenes and user intentions. 
To support this task, we also develop a dedicated data curation pipeline and construct CT-200K, a large-scale VLC dataset containing dynamic videos with over 47 million frames. 
The pipeline addresses the scarcity of reliable camera-motion annotations through a carefully designed collaboration of task-specific and general-purpose models. 
Extensive experiments on CameraBench~\cite{lin2025towards} show that CT-1 predicts camera trajectories that are structurally grounded and semantically aligned with user intent. When coupled with a controllable video generator, our method significantly outperforms existing state-of-the-art approaches, improving camera control success rate by 25.7\%.
In summary, our main contributions are as follows: \looseness=-1

\begin{itemize}[leftmargin=*, noitemsep, topsep=10pt]
    \item We introduce a novel Vision–Language–Camera (VLC) model for camera-controllable video generation that jointly reasons over visual observations and semantic instructions to infer spatially aware, temporally coherent camera trajectories.
    
    \item We propose CT-1 (Camera Transformer~1), a Transformer-based instantiation of the VLC model that injects camera-context tokens from vision–language modules into a Diffusion Transformer, enabling distributional modeling of camera trajectories with Wavelet-based Regularization Loss.
    
    \item To support the training of the CT-1 model, we carefully design a data curation pipeline and build a large-scale dataset, CT-200K, which contains over 47 million frames across multiple scenarios.
    
    \item To the best of our knowledge, this work represents an early attempt to integrate camera trajectory estimation with multimodal modeling for camera-controllable video generation. Experimental results demonstrate that our framework achieves improved camera-controllable video synthesis and validate the effectiveness of the CT-1 model.
\end{itemize}

%% file: section/related.tex
\section{Related Works}

\subsection{Video Generation}
Recent advances in diffusion models based on U-Net and Diffusion Transformer (DiT)~\cite{peebles2023scalable} architectures have significantly improved video generation~\cite{chen2024videocrafter2,zhao2025magdiff,yang2024cogvideox,wan2025wan,hacohen2024ltx,zhao2026lstd,yu2025context,zhengrefocuseraser}. U-Net–based approaches employ an encoder–decoder structure that offers efficient inference and architectural flexibility, but their limited model capacity constrains performance gains under large-scale scaling. As a result, recent methods, such as CogVideoX~\cite{yang2024cogvideox} and Wan model~\cite{wan2025wan}, increasingly adopt DiT as the core building block, leveraging stacked Transformer layers to jointly model spatial and temporal dependencies, with scalable depth that better supports large-scale training.

\subsection{Camera-Controllable Video Generation}
Controllable video generation enhances general video diffusion models by introducing additional conditioning signals, typically through condition concatenation~\cite{zhao2025magdiff} and cross-attention~\cite{wang2025uniadapter}.
For camera-controllable video generation, existing methods often require explicitly specified camera parameters to be directly injected into the model (\textit{e.g.}, MotionCtrl~\cite{wang2024motionctrl}, CameraCtrl~\cite{he2025cameractrl}), aiming to induce corresponding view changes in the videos. However, due to the lack of explicit modeling of semantic and spatial alignment between the camera and visual content, these approaches frequently produce spatially inconsistent or visually implausible results.

\subsection{Camera Trajectory Estimation}
Camera trajectories describe the pose variations of a camera during the capture process and are jointly determined by camera intrinsics and extrinsics. 
Existing methods, such as CameraCtrl~\cite{he2025cameractrl} and MotionCtrl~\cite{wang2024motionctrl}, typically constrain video generation by explicitly injecting predefined or video-extracted camera extrinsics, thereby controlling the resulting visual changes. 
However, how to automatically generate plausible camera trajectories from high-level semantic information remains largely underexplored. 
Although ChatCam~\cite{liu2024chatcam} synthesizes camera motion solely from textual prompts, its generated trajectories lack awareness of the underlying visual content.
Motivated by these observations, we introduce CT-1, which aims to model and generate camera trajectories under the joint visual content and language instructions.

%% file: section/method.tex
\begin{figure*}[t!]
    \centering
    \captionsetup{type=figure}
    \includegraphics[width=1.0\linewidth]{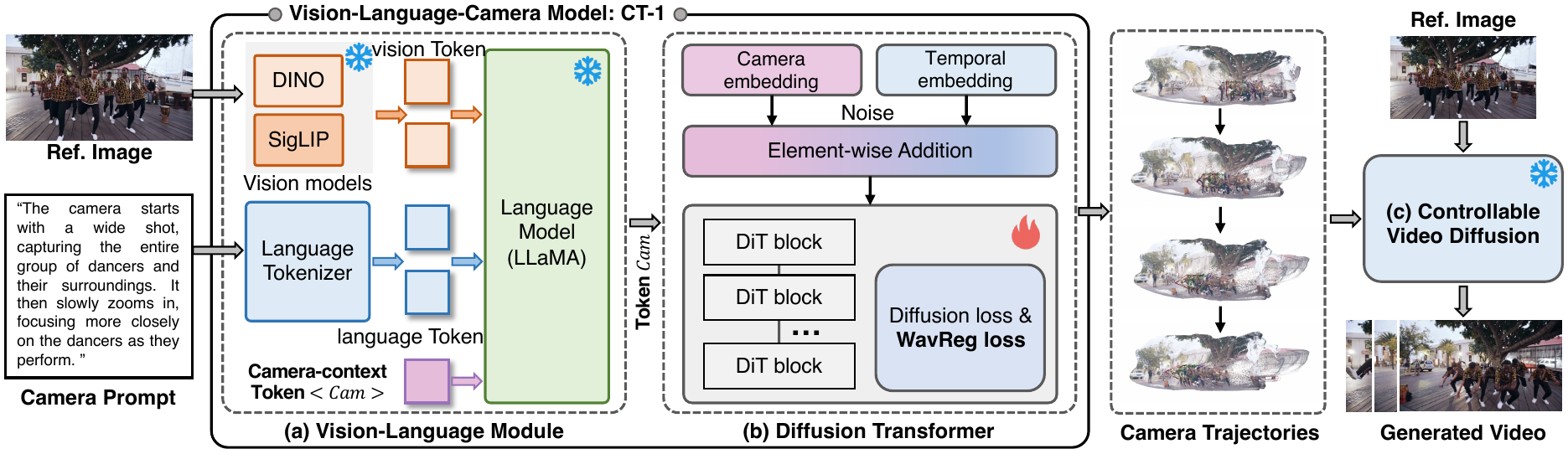}
    \caption{Overview of the proposed camera-controllable video generation framework based on the CT-1 model, which includes (a) a vision–language module for semantic embedding, (b) a Diffusion Transformer module for modeling camera trajectory distributions, and (c) controllable generation models that synthesize videos conditioned on the trajectories.}
    \label{fig:framework}
\end{figure*}

\section{Method}

In this section, we introduce a camera-controllable video generation framework with a novel VLC model, termed CT-1, as shown in Fig.~\ref{fig:framework}.
We begin by describing the formulation of the VLC model, highlighting how it enables precise camera prediction.
Then, we present the methodology of our comprehensive framework within the CT-1 to achieve camera control, including vision-language modules, camera transformer module, and the video generation module.
Finally, we present CT-200K, a dataset collected for training.

\subsection{Problem Formulation}
\label{appendix:problem_formulation}
As illustrated in Fig.~\ref{fig:intro} (a), the VLC model aims to learn spatial reasoning knowledge and further predict camera trajectories, which acts as a mapping function $p_{\theta}$ from semantic input to camera motion output.
It determines how the camera moves over time $t$ from an initial visual observation $v$ and textual instruction $\ell$.
The VLC model predicts a camera trajectory $K_{1:T} = \{K_t\}_{t=1}^T$, where each camera pose $K_t \in SE(3)$ is parameterized by a rotation matrix $R_t \in SO(3)$ and a translation vector $\mathbf{p}_t \in \mathbb{R}^3$, formally as:

\begin{equation}
p_\theta(K_{1:T}\mid v,\ell).
\end{equation}

Unlike token-based sequence generation, each predicted pose $K_t$ lies on the $SE(3)$ manifold, which imposes strict geometric constraints on camera motion.
We emphasize that the camera trajectories estimated by the VLC model exhibit the following properties:
1) \textit{Semantic Alignment}, where the trajectories conform to the motion semantics of the vision–language instructions and form continuous curves on $SE(3)$;
2) \textit{Temporal Continuity}, where the trajectories are Lipschitz continuous with respect to time on $SE(3)$, which ensures smooth and stable camera motion; and
3) \textit{Non-Uniqueness}, where multiple semantically valid trajectories may exist for the same instruction.
Further details are in Appendix~\ref{appendix:vlc_explanation}.

\subsection{Framework with CT-1}

\subsubsection{Vision and language module}
\textbf{Vision module.} Our vision module adopts a dual-branch architecture that encodes raw image inputs into a unified set of visual perceptual tokens. Specifically, it consists of two parallel vision encoders, DINOv2~\cite{oquab2023dinov2} and SigLIP~\cite{zhai2023sigmoid}. These two encoders capture complementary visual cues, ranging from fine-grained local details to high-level semantic information, thereby enhancing the expressiveness of the visual representation.
Given the image $v$, we feed it into both visual encoders, producing downsampled feature maps. The features from the two branches are then concatenated along the channel dimension and projected into a shared space via a linear projection layer. 

\noindent \textbf{Language module.}
The language module adopts a large language model, which plays a central role in jointly modeling visual observations and language instructions, enabling multimodal fusion and high-level cognitive reasoning, shown in Fig.~\ref{fig:framework} (a). We employ LLaMA-2~\cite{touvron2023llama} as the language backbone, providing strong linguistic modeling capacity for cross-modal understanding.
At the input stage, a language instruction $\ell$ is first tokenized using the tokenizer into a sequence of linguistic tokens $\mathcal{T} = {\ell_1, \ell_2, \ldots, \ell_{N_T}}.$
These linguistic tokens, together with the visual perceptual tokens $\mathcal{V}$ and a learnable camera-context token \texttt{<CAM>}, are concatenated into a unified sequence and processed via a causal attention mechanism, enabling deep multimodal integration.
At the output stage, the hidden representation $f_t^{cam}$ corresponding to the \texttt{<CAM>} serves as a compact global feature that aggregates visual and linguistic information at the current timestep. This representation is then used as a conditioning signal for the subsequent camera transformer structure to interpret and derive the desired trajectories.

\subsubsection{Camera transformer}
In previous vision-language-based numerical regression settings, such as directly predicting object coordinates~\cite{chen2023large} or bounding boxes~\cite{du2022learning} from images, the output tokens can be fed into a regression head to learn deterministic numerical outputs.
However, our VLC model requires predicting camera trajectories in semantic space, where the objective is to model a distribution over plausible camera motions rather than to infer a single fixed numerical solution. 
Therefore, we adopt a diffusion model based on Diffusion Transformers (DiTs) to learn the distribution of trajectories and propose a Wavelet-based Regularization Loss in the frequency domain, shown in Fig.~\ref{fig:framework} (b).

\noindent \textbf{Basic structure.} 
For diffusion-based modeling, we represent the trajectory as a sequence of $C$-dimensional Euclidean pose parameters, yielding $\mathrm{K} \in \mathbb{R}^{T\times C}$, while retaining the $SE(3)$ interpretation in the problem formulation.
We define the forward diffusion process at diffusion step $s$ as:

\begin{equation}
\mathrm{K}^{(s)} = \sqrt{\bar{\alpha}_s}\,\mathrm{K} + \sqrt{1-\bar{\alpha}_s}\,\epsilon,
\qquad
\epsilon \sim \mathcal{N}(0, I),
\end{equation}

\noindent where $\mathrm{K}^{(s)} \in \mathbb{R}^{T \times C}$ denotes the noisy trajectory and $\bar{\alpha}_s$ is the cumulative noise schedule. We first project $\mathrm{K}^{(s)}$ into a latent space. The diffusion step $s$ is encoded as a timestep embedding $e_s \in \mathbb{R}^{1 \times D}$. In parallel, a camera-context token \texttt{<CAM>} is mapped to a learnable embedding $e_{\mathrm{cam}} \in \mathbb{R}^{1 \times D}$. These two embeddings are combined by element-wise addition to form a conditional token $c = e_s + e_{\mathrm{cam}}$, which is prepended to the embedded trajectory sequence. The resulting token sequence is further augmented with learnable positional embeddings and processed by a stack of DiT blocks. Finally, a linear projection maps the latent outputs back to the trajectory space, yielding the predicted clean trajectory $\hat{\mathrm{K}} \in \mathbb{R}^{T\times C}$.

\begin{figure*}[t]
    \centering
    \captionsetup{type=figure}
    \includegraphics[width=1.0\linewidth]{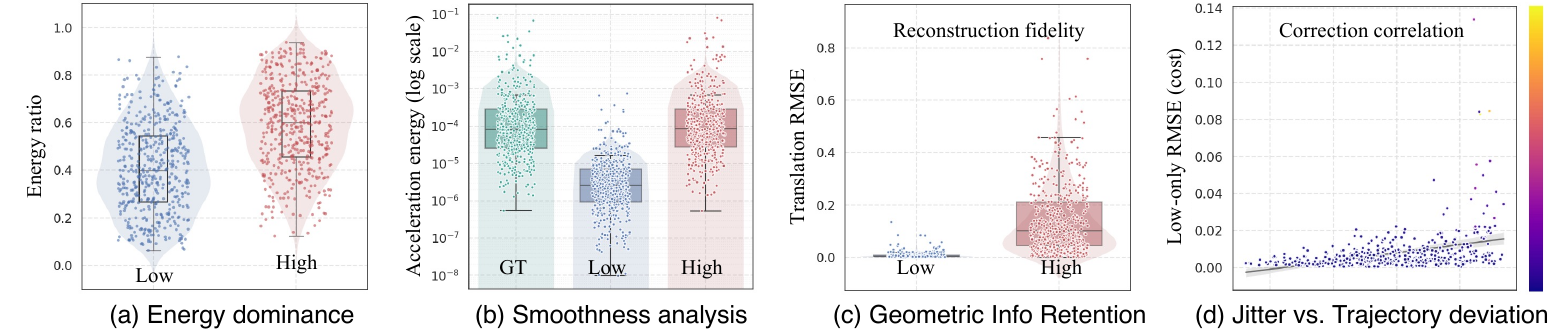}
    \caption{Wavelet analysis of camera trajectories over 1,000 examples, including energy dominance, smoothness, geometric retention, and the trade-off between jitter and trajectory deviation across low- and high-frequency components.}
    \label{fig:wavelet_analysis}
\end{figure*}

\noindent \textbf{Wavelet-based Regularization Loss.}
Moreover, we demonstrate that the camera trajectories are naturally viewed as a multi-scale structure, including low- and high-frequency components. To prove this, we apply a 1D orthonormal Haar Discrete Wavelet Transform (DWT) to decompose camera trajectories into low-frequency and high-frequency signals. Shown in Fig.~\ref{fig:wavelet_analysis}, we analyze the trajectories from three perspectives: energy distribution, temporal smoothness, and reconstruction error.
More individual visualizations are conducted in Appendix~\ref{appendix:wavelet_decomposition}.
From these results, we draw the following conclusions: 
1) \textit{Distinct motion roles}: low-frequency components capture global and smooth motion, while high-frequency components model local, rapid perturbations.
2) \textit{Structural fidelity}: low-frequency components reconstruct camera translation and rotation with negligible error, encoding the principal trajectory structure.
3) \textit{Global dominance}: low-frequency signals dominate global camera motion, preserving overall displacement even without high-frequency components.

These observations highlight the importance of frequency-aware regularization for camera trajectory modeling. Accordingly, we introduce a Wavelet-based Regularization Loss (WavReg) to encourage stable and smooth camera trajectories.
Beginning with the predicted clean trajectory $\hat{\mathrm{K}} \in \mathbb{R}^{T\times C}$, reconstructed from the model output at diffusion step $s$, we apply the DWT along the temporal dimension, one level of decomposition as:

\begin{equation}
a_l^{(1)} = \frac{\hat{\mathrm{K}}[2l-1,:] + \hat{\mathrm{K}}[2l,:]}{\sqrt{2}},
\qquad
d_l^{(1)} = \frac{\hat{\mathrm{K}}[2l-1,:] - \hat{\mathrm{K}}[2l,:]}{\sqrt{2}},
\end{equation}

\noindent where $\hat{\mathrm{K}}[t,:] \in \mathbb{R}^{C}$ denotes the trajectory vector at time step $t$, $a_l^{(1)}$ is the low-frequency approximation, $d_l^{(1)}$ is the high-frequency detail, and $l$ indexes temporal pairs at the first decomposition level. Accordingly, both $a_l^{(1)}$ and $d_l^{(1)}$ are $C$-dimensional vectors.
We apply this recursively for multiple $L$ levels, yielding:

\begin{equation}
W(\hat{\mathrm{K}}) =
\left(
a_L(\hat{\mathrm{K}}),\,
d_L(\hat{\mathrm{K}}),\,
d_{L-1}(\hat{\mathrm{K}}),\,
\ldots,\,
d_1(\hat{\mathrm{K}})
\right),
\end{equation}

\noindent where $a_L$ captures the global, semantic camera motion trend, $d_l$ captures increasingly fine-grained temporal disturbances. We adopt the orthonormal Haar basis (scaling factor $1/\sqrt{2}$), which ensures:

\begin{equation}
\|\mathrm{x}\|_2^2
=
\|a_L(\mathrm{x})\|_2^2
+
\sum_{l=1}^{L}\|d_l(\mathrm{x})\|_2^2.
\end{equation}

Here, $\mathrm{x}$ denotes an arbitrary trajectory sequence. This property makes coefficients across different scales numerically comparable.
When performing this decomposition, we pad the temporal dimension to be divisible by $2^L$, apply reflect padding to mitigate boundary-induced bias, and recursively compute the low- and high-frequency components.
Then, we define the WavReg loss as:

\begin{equation}
\label{eq:12}
\mathcal{L}_{\mathrm{wav}}
=
\lambda_a \left\| a_L(\hat{\mathrm{K}}) - a_L(\mathrm{K}) \right\|_1
+
\sum_{l=1}^{L}
\lambda_{d_l}
\left\| d_l(\hat{\mathrm{K}}) - d_l(\mathrm{K}) \right\|_1.
\end{equation}

\noindent In Eq.~\ref{eq:12}, $\hat{\mathrm{K}}$ denotes the clean trajectory reconstructed from the denoiser output, where $a_L(\cdot)$, $d_l(\cdot)$, and $\lambda_a, \lambda_{d_l}$ are the final low-frequency component, the coefficients at scale $l$, and scale-dependent weights, respectively.
Motivated by observations that camera motion energy is dominated by low-frequency signals, we assign:

\begin{equation}
\label{eq:lamda}
\lambda_a > \lambda_{d_L} > \dots > \lambda_{d_1},
\end{equation}

\noindent where $\lambda_a$ and $\lambda_{d_l}$ are scale-dependent weights, with larger weights assigned to lower-frequency components.
The full training objective combines the standard diffusion loss and WavReg loss with weight $\beta$. The full training objective is:

\begin{equation}
\label{eq:total_loss_method}
\mathcal{L} = \mathcal{L}_{\text{diff}} + \beta\,\mathcal{L}_{\text{wav}}.
\end{equation}

\subsubsection{Camera-controllable Video Generation}
Given a reference image $v$, a description $\ell$, and the trajectory $K_{1:T}$ predicted by CT-1, we generate a video sequence $x_{1:T}$ by conditioning a controllable video diffusion model on the estimated camera trajectory. We model the conditional video distribution as:

\begin{equation}
\label{eq:video_generation}
p_\Phi(x_{1:T}\mid v,l,K_{1:T}).
\end{equation}

In our framework, CT-1 and the video generator are decoupled in a modular two-stage pipeline. CT-1 is responsible for predicting a semantically aligned and temporally coherent camera trajectory, while the video diffusion model uses this trajectory as the motion control signal for synthesis.
In the main experiments, we instantiate the video generator using CameraNoise~\cite{cameranoise}. The predicted trajectory is then converted into the corresponding camera control representation and injected into the diffusion process to guide video generation.
This design enables CT-1 to transfer spatial reasoning knowledge to the downstream video generator without modifying the core architecture of the backbone. Owing to its modularity, the same trajectory interface can also be applied to other controllable video diffusion models, such as CameraCtrl~\cite{he2025cameractrl} and MotionCtrl~\cite{wang2024motionctrl}.

\begin{figure*}[t]
    \centering
    \captionsetup{type=figure}
    \includegraphics[width=1.0\linewidth]{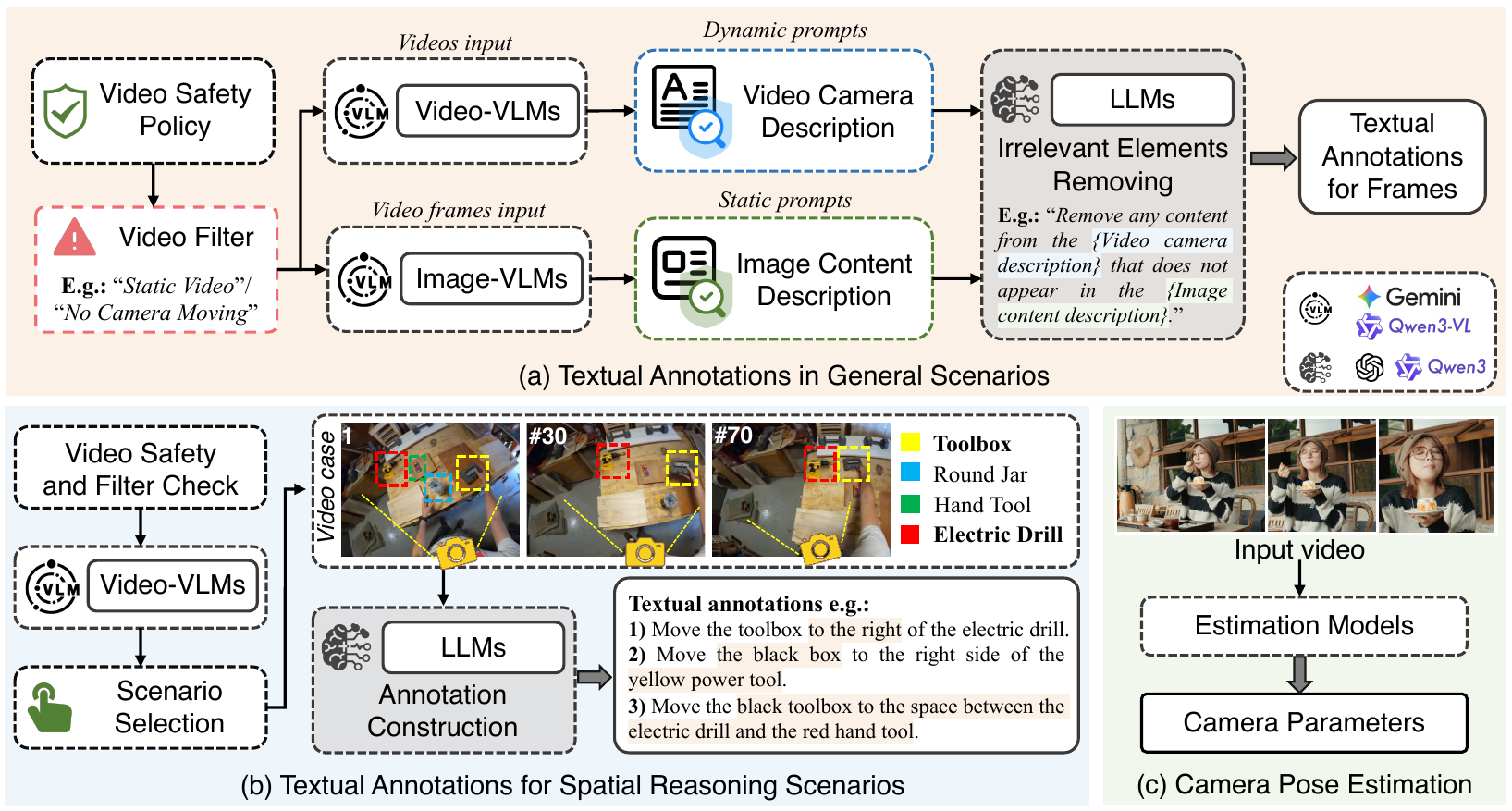}
    \caption{Overview of the CT-200K data curation pipeline, designed to address missing or ambiguous camera-motion annotations in existing datasets and to support annotation for reasoning scenarios.}
    \label{fig:data_curation_pipeline}
\end{figure*}

\subsection{CT-200K: VLC Dataset Construction}
The foundation of CT-200K is a diverse and contextually rich video corpus. 
Since existing datasets lack camera motion descriptions and corresponding parameters, we leverage vision-rich video sequences together with the multimodal capabilities of LLMs and MLLMs to construct our dataset, shown in Fig.~\ref{fig:data_curation_pipeline} (a) and (b).
Because the VGGT~\cite{wang2025vggt} model achieves a camera pose estimation accuracy of up to 93.5\% in both indoor and outdoor scenes, we use it to annotate unlabeled videos, shown in Fig.~\ref{fig:data_curation_pipeline} (c).
The generated description candidates are carefully reviewed by human annotators to ensure the diversity and quality of the dataset.
Specifically, two scenarios are considered: general and reasoning video scenarios.

\noindent \textbf{General video scenarios.}
We begin with the Pexels-400K dataset~\cite{pexels} and DynPose-100K~\cite{rockwell2025dynamic}, selected for their high-quality and diversity.
Shown in Fig.~\ref{fig:data_curation_pipeline} (a), we first filter videos using inter-frame optical flow differences, removing clips with overly fast, negligible, or static camera motion.
The remaining videos and their corresponding frames are then processed by a Video-VLM~\cite{bai2025qwen2} to generate video camera descriptions $\mathrm{QA_V}$ that capture camera motion patterns, and an Image-VLM~\cite{bai2025qwen2} to produce image content descriptions $\mathrm{QA_I}$ that reflect per-frame visual semantics.
Next, we employ an LLM~\cite{yang2025qwen3} with a structured instruction ``Remove any content in the \texttt{<$\mathrm{QA_V}$>} that does not appear in the \texttt{<$\mathrm{QA_I}$>}'' to refine the camera descriptions.
This choice is motivated by the fact that Video-VLMs often introduce cross-temporal or anticipatory information, which would break vision–language consistency and cause a mismatch when an image and its trajectory are used.

\noindent \textbf{Reasoning video scenarios.}
To enhance visual reasoning, we build reasoning scenarios from egocentric videos, where a camera wearer manipulates objects in a fixed area, and the head-mounted camera captures the resulting motion.
As illustrated in Fig.~\ref{fig:data_curation_pipeline} (b), the scene contains a ``toolbox'', a ``round jar'', a ``hand tool'', and an ``electric drill''. When the operator moves the ``toolbox'' (yellow dashed box) from the lower-right corner of the table to the upper-left corner, the camera trajectory undergoes a corresponding displacement. We construct such pairs that contain implicit camera motion, \textit{e.g.,} ``move the toolbox to the right of the electric drill''. 
Based on the EgoSchema~\cite{mangalam2023egoschema}, EgoMe~\cite{qiu2025egome}, and Ego-ST~\cite{wu2025st} datasets, we introduce additional processing steps on top of the pipeline for general scenarios. Specifically, we use descriptions generated by a Video-VLM to identify and filter videos that correspond to the desired reasoning scenarios. We then employ LLMs to reconstruct the corresponding instruction annotations from $\mathrm{QA_I}$.
This design encourages the model to infer and reason about spatial relationships and camera motion jointly, thereby strengthening its spatial understanding and reasoning capability.

%% file: section/results.tex
\section{Experiments}

\subsection{Implementation Details}

\textbf{Evaluation metrics.}
Because camera trajectory prediction is inherently non-unique, we adopt Success Rate as the primary metric for camera controllability. Following evaluation protocols~\cite{kim2024openvla, kim2025fine} used in motion-planning-style tasks, we assess whether a generated camera trajectory is both semantically consistent with the instruction and physically plausible. Specifically, two human experts independently evaluate each result based on the reference image, the camera-motion description, and the generated video. A sample is counted as successful only when both experts agree that the generated camera motion correctly follows the intended instruction. We additionally report VBench~\cite{huang2024vbench} to measure the visual quality of the generated videos.

\noindent \textbf{Evaluation datasets.}
We evaluate our method on CameraBench100, a 100-sample subset of CameraBench~\cite{lin2025towards} that covers six representative camera-motion settings and diverse scenes, including daily-life, cinematic, and virtual environments.
For qualitative evaluation, we further test on RealEstate10K~\cite{zhou2018stereo}, MultiCamVideo~\cite{bai2025recammaster}, and DrivingDoJo~\cite{wang2024drivingdojo} datasets.

\noindent \textbf{Video backbones.}
Based on the estimated trajectories, we use CameraNoise~\cite{cameranoise} with Wan2.1~\cite{wan2025wan} as the training base model in our main experiments and ablations. We further report results with CameraCtrl~\cite{he2025cameractrl} and MotionCtrl~\cite{wang2024motionctrl} in Appendix~\ref{appendix:applying2others}.

\noindent \textbf{Baselines.}
We compare CT-1 and video generation performance against state-of-the-art camera-controllable diffusion models, including MotionCtrl~\cite{wang2024motionctrl} and CameraCtrl~\cite{he2025cameractrl}, as well as I2V models, including CogVideoX~\cite{yang2024cogvideox}, LTX-Video~\cite{hacohen2024ltx}, and Wan 2.1/2.2~\cite{wan2025wan}. More implementation details are in Appendix~\ref{appendix:implementation_details}.

\begin{table*}[t!]
\centering
\small
\caption{Zero-shot quantitative results on CameraBench100 for camera trajectory estimation across six typical tasks using the Success Rate metric. PE denotes Prompt Extension for fair comparison, and AR denotes autoregressive trajectory generation.}
\resizebox{1.0\linewidth}{!}{
\begin{tabular}{l|l|cccc|cc|c}
\toprule
\multirow{2}{*}{\textbf{Inference Type}} &
  \multirow{2}{*}{\textbf{Method}} &
  \multirow{2}{*}{\textbf{Dolly-in}} & \textbf{Pan Left} & \textbf{Truck Left} & \textbf{Tilt Down} & \textbf{Regular} & \textbf{Complex} & \multirow{2}{*}{\emph{\textbf{Average}}$\:\uparrow$} \\ 
  & & & \textbf{or Right} & \textbf{or Right} & \textbf{or Up} & \textbf{Speed} & \textbf{Motion} & \\
  \midrule
\multirow{2}{*}{\begin{tabular}[l]{@{}l@{}} Trajectory-input-based using VLM
\end{tabular}}
             & MotionCtrl~\cite{wang2024motionctrl}  & 24.0     & 31.1     & 18.5     & 27.5    & 30.2 & 21.8 & 25.5  \\
             & CameraCtrl~\cite{he2025cameractrl}   & 29.0     & 36.1    & 22.6  &  31.3 & 34.7 & 27.0 & 30.1  \\ 
             \midrule
\multirow{2}{*}{\begin{tabular}[l]{@{}l@{}} Trajectory-input-based using AR
\end{tabular}}
             & MotionCtrl~\cite{wang2024motionctrl}  & 18.0     &  24.4    & 14.3     & 22.5    & 24.9 & 17.1 & 20.2  \\
             & CameraCtrl~\cite{he2025cameractrl}   & 22.0     & 28.9    & 16.7  & 25.0 & 28.1 & 20.6 & 23.6  \\ 
             \midrule
\multirow{4}{*}{\begin{tabular}[l]{@{}l@{}} Prompt-input-based w/ PE \end{tabular}}
             & LTX-Video~\cite{hacohen2024ltx}      & 34.0     & 58.3     & 28.6     & 50.0    & 56.2 & 45.7 & 45.5  \\ 
             & CogVideoX~\cite{yang2024cogvideox}   & 28.0     & 41.7    & 21.4  &  37.5 & 48.3 & 32.6 & 34.9  \\ 
             & Wan2.1~\cite{wan2025wan}      & 40.0     & 58.3     & 35.7     & 50.0     & 64.0 & 59.8 & 51.3  \\
             & Wan2.2~\cite{wan2025wan}      & \underline{68.0}     & \underline{63.9}     & 50.0     & \underline{62.5}     & \underline{76.4} & \underline{68.5} & \underline{64.9}  \\
             \midrule
\multirow{5}{*}{\begin{tabular}[l]{@{}l@{}} Prompt-input-based w/o PE \end{tabular}}
             & LTX-Video~\cite{hacohen2024ltx}  & 38.0    & 55.6    & 28.6     &  50.0    & 53.9 & 48.9 & 45.8   \\ 
             & CogVideoX~\cite{yang2024cogvideox}  & 30.0  & 38.9     & 21.4     &  37.5   &  47.2  & 34.8 & 35.0  \\ 
             & Wan2.1~\cite{wan2025wan}  & 44.0     & 55.6     & 42.9     &  50.0   &  61.8  & 54.3 & 51.4  \\
             & Wan2.2~\cite{wan2025wan}  & 64.0     & 58.3     & \underline{57.1}     &  \underline{62.5}   &  70.8  & 59.8 & 62.1  \\
             \cmidrule{2-9}
             & \textbf{Ours}  & \textbf{88.0}     & \textbf{86.1}     & \textbf{71.4}     & \textbf{75.0}     & \textbf{87.6}    & \textbf{81.5} & \textbf{81.6}  \\    
\bottomrule
\end{tabular}}
\label{tab:success_rate}
\end{table*}

\subsection{Comparisons with Other Methods}

\textbf{Quantitative comparison of camera control.}
Our model takes a camera-motion description and a reference image as input. To comprehensively evaluate its camera control capability, we compare with two categories of methods: \textit{trajectory-input-based models}, which estimate camera trajectories using either VLMs or AR models and then guide video generation, and \textit{prompt-input-based models}, which directly use camera-motion descriptions with or without Prompt Extension (PE). For the former, we use Qwen3-VL~\cite{bai2025qwen3} and DVGFormer~\cite{hou2024learning} as representative VLM- and AR-based trajectory estimators, respectively. For the latter, we evaluate both PE-enhanced (w/ PE) and original prompts (w/o PE). Our method belongs to the prompt-input-based setting without PE, as it directly generates videos from the image-text pairs.

As shown in Table~\ref{tab:success_rate}, our method achieves the best average camera control accuracy among all compared methods. It improves the average success rate by \textbf{171.1\%} over the best VLM-based trajectory-input model, by \textbf{245.8\%} over the best AR-based trajectory-input model, by \textbf{25.7\%} over the best prompt-input-based model with PE, and by \textbf{31.4\%} over the best prompt-input-based model without PE. Our method also shows clear advantages in challenging scenarios such as \textit{Truck Left or Right} and \textit{Complex Motion}, further proving strong controllability in camera estimation and video generation.

\begin{table}[h]
\centering
\small
\caption{Comparison of generated video quality with baseline methods using VBench. Results w/o PE are in gray.}
\resizebox{0.6\linewidth}{!}{
\begin{tabular}{l|cccc}
\toprule
  \multirow{2}{*}{Method} &
  Aesthetic & Imaging & Motion & Dynamic \\ 
  & Quality$\:\uparrow$ & Quality$\:\uparrow$ & Smoothness & Degree \\
  \midrule
             \rowcolor{lightgray!20}
             LTX-Video~\cite{hacohen2024ltx}      & 0.535     & 0.628    & 0.972     & \textbf{0.981} \\ 
             \rowcolor{lightgray!20}
             CogVideoX~\cite{yang2024cogvideox}   & 0.540     & 0.681     & \underline{0.985}     & 0.445 \\  
             \rowcolor{lightgray!20}
             Wan2.1~\cite{wan2025wan}     & 0.565     & 0.685     & 0.977     & 0.505 \\ 
             \rowcolor{lightgray!20}
             Wan2.2~\cite{wan2025wan}      & 0.548     & 0.665     & 0.970     & 0.810 \\ 
             \midrule
             LTX-Video~\cite{hacohen2024ltx}  & 0.542  & 0.634    & 0.973     &  \underline{0.980}  \\ 
             CogVideoX~\cite{yang2024cogvideox}  & 0.536  & 0.675    & 0.981     &  0.450   \\ 
             Wan2.1~\cite{wan2025wan}     & \underline{0.571}   & \underline{0.691}  & 0.980      &  0.510   \\
             Wan2.2~\cite{wan2025wan}     & 0.549  & 0.671    & 0.977     &  0.808    \\
             \midrule
             \textbf{Ours}  & \textbf{0.585}  & \textbf{0.709}  & \textbf{0.990}  & 0.830 \\
\bottomrule
\end{tabular}}
\label{tab:vbench}
\end{table}

\noindent \textbf{Quantitative comparison of video quality.}
We employ VBench to evaluate the quality of videos generated by different models. The results in Table~\ref{tab:vbench} indicate that, across VBench metrics, the video quality produced by our method is comparable to that of existing state-of-the-art video backbones. 
This suggests that CT-1 estimates reasonable camera trajectories in both motion and structure, while preserving temporal consistency in the generated videos.

\begin{figure*}[t!]
    \centering
    \captionsetup{type=figure}
    \includegraphics[width=1.0\linewidth]{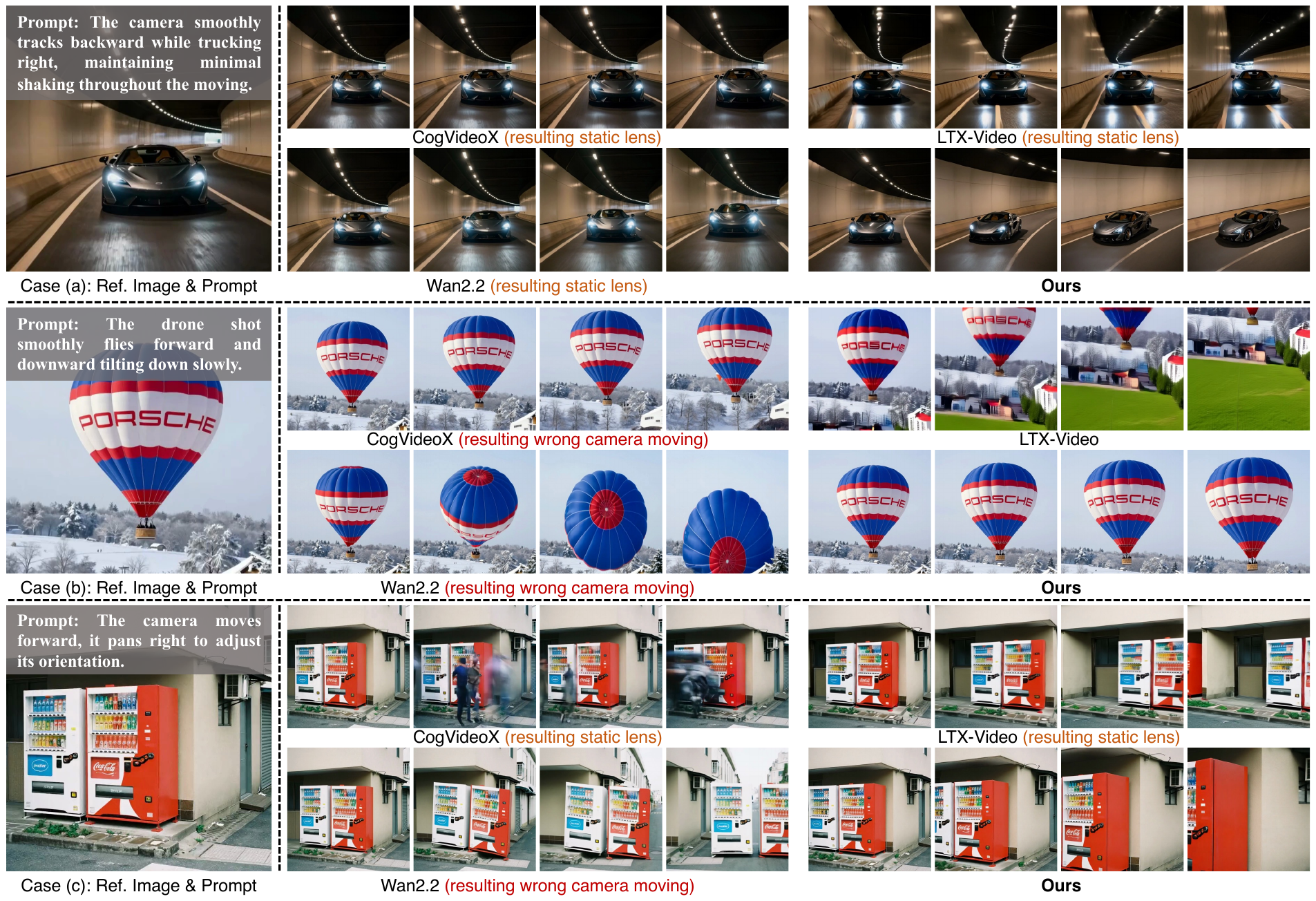}
    \caption{Qualitative comparison in out-of-distribution (OOD) scenarios with CogVideoX~\cite{yang2024cogvideox}, LTX-Video~\cite{hacohen2024ltx}, and Wan2.2~\cite{wan2025wan}.}
    \label{fig:qualitative_result}
    \vspace{-10pt}
\end{figure*}

\noindent \textbf{Qualitative comparison.}
In Fig.~\ref{fig:qualitative_result}, we compare CT-1 with existing methods on dynamic and OOD scenarios. 
The results expose a key limitation of current backbones: low sensitivity to camera-motion text. For example, prompts like ``tracks backward while trucking right'' often yield nearly static camera motion.
\textbf{We provide more extensive visualizations and analyses in the appendix} to further validate the effectiveness of our approach, including \textbf{1)} trajectory visualizations under different description granularities in Appendix~\ref{appendix:camera_trajectories}; \textbf{2)} cross-condition trajectory generation with multiple images or texts in Appendix~\ref{appendix:cross_validation}; \textbf{3)} results under challenging scenarios in Appendix~\ref{appendix:challenging_scenarios}; \textbf{4)} applying CT-1 on other controllable video diffusion models in Appendix~\ref{appendix:applying2others}; \textbf{5)} results in deep-reasoning-based scenarios in Appendix~\ref{appendix:deep_reasoning}; and \textbf{6)} application to dynamic driving scenarios in Appendix~\ref{appendix:driving_scenarios}.

\begin{table}[h]
\centering
\small
\caption{Ablation of camera control and video generation performance across different CT-1 scales.}
\resizebox{0.6\linewidth}{!}{
\begin{tabular}{l|c|cc|cc}
\toprule
  \multirow{2}{*}{Method} & \multirow{2}{*}{Param.} &
  Imaging & Dynamic & Regular & Complex \\ 
  & & Quality & Degree & Speed & Motion  \\
  \midrule
             CT-1 (Base)  & 33M  & 0.623   & 0.674     & 72.3     & 64.8   \\
             CT-1 (Large)  & 130M & \underline{0.685}  & \underline{0.769}    & \underline{82.7}    &  \underline{76.4}    \\
             \midrule
             CT-1 (Huge) & 458M & \textbf{0.708}  & \textbf{0.830}  & \textbf{87.6}  & \textbf{81.5} \\
\bottomrule
\end{tabular}}
\label{tab:scaling_model_size}
\end{table}

\subsection{Ablation Study}
\textbf{Impact of model scaling.}
We train three CT-1 models of different scales based on the DiT architecture, and systematically explore model scaling under the \textit{Base, Large}, and \textit{Huge} configurations. These DiT models contain 33M, 130M, and 458M parameters, respectively. Table~\ref{tab:scaling_model_size} summarizes the performance comparison across different model scales, evaluated using VBench as well as success rates under the ``Regular Motion'' and ``Complex Motion'' settings. The results show that overall performance consistently improves with increasing model size, validating that CT-1 exhibits scaling behavior similar to that of DiT models and suggesting its strong potential for further performance gains and broader applicability at larger scales.

\begin{table}[ht]
\centering
\small
\caption{Impact of the WavReg loss and the parameter $\beta$ on the performance of the CT-1 model.}
\resizebox{0.6\linewidth}{!}{
\begin{tabular}{l|cc|cc}
\toprule
  \multirow{2}{*}{Method} &
  Imaging & Dynamic & Regular & Complex \\ 
  & Quality & Degree & Speed & Motion  \\
  \midrule
            \rowcolor{lightgray!20}
             WavReg / $\beta=0$     & 0.681  & 0.802    &  82.0  &   76.1   \\
             \midrule
             WavReg / $\beta=1.0$     & 0.678   & 0.799     &   80.9    &  77.2   \\
             WavReg / $\beta=0.5$     & 0.682   & 0.808     &   82.0    &  79.3   \\
             WavReg / $\beta=0.3$     & \underline{0.702}   & \underline{0.824}     &   83.1    &  \underline{80.4}   \\
             WavReg / $\beta=0.05$     & 0.697   & 0.816     & \underline{84.3}     & 79.3  \\
             \midrule
             WavReg / $\beta=0.1$  & \textbf{0.708}  & \textbf{0.830}  & \textbf{87.6}  & \textbf{81.5} \\
\bottomrule
\end{tabular}}
\label{tab:wavreg}
\end{table}

\noindent \textbf{Role of the WavReg loss.}
In CT-1, we introduce WavReg, a frequency-domain regularization term, which is combined with the diffusion loss via a weighted formulation as defined in Eq.~(\ref{eq:total_loss_method}). To analyze the effect of WavReg, we conduct an ablation study on the weighting coefficient $\beta$, with results reported in Table~\ref{tab:wavreg}. When $\beta = 0$, WavReg is entirely disabled and does not contribute to training. As $\beta$ is gradually increased, the overall model performance first improves and then stabilizes, achieving the best results at $\beta = 0.1$. More theoretical proofs are provided in Appendix~\ref{appendix:effect_of_beta}.

\begin{table}[ht]
\centering
\small
\caption{Ablation analysis under the settings of the same image with different prompts (Image $\rightarrow$ Prompts) and the same prompt with different images (Prompt $\rightarrow$ Images).}
\resizebox{0.6\linewidth}{!}{
\begin{tabular}{l|cc|cc}
\toprule
  \multirow{2}{*}{Method} &
  Imaging & Dynamic & Regular & Complex \\ 
  & Quality & Degree & Speed & Motion  \\
  \midrule
              Prompt $\rightarrow$ Images  & 0.663  & 0.780   & 82.0  &  75.0  \\
              \rowcolor{lightgray!20}
              Image $\rightarrow$ Prompts  & 0.725  & 0.810   & 88.8   & 85.9   \\
\bottomrule
\end{tabular}}
\label{tab:camera_learning}
\end{table}

\noindent \textbf{Learning dynamics of camera trajectories.}
Since CT-1 does not use a VLM with a regression head, it learns a distribution of camera trajectories using a diffusion process. We evaluate it via cross-validation under two settings: the same prompt across different scenes, and the same scene with different prompts.
Specifically, we select 10 general camera descriptions and 10 representative scenes from CameraBench100. Each description is tested on ten scenes, and each scene is tested with ten descriptions.
Under this experimental setup, we report the results for different combinations in Table~\ref{tab:camera_learning}. The results demonstrate that, even under repeated and cross-conditioned evaluation, CT-1 maintains stable camera trajectory generation under repeated and cross-conditioned evaluation.

\begin{table}[ht]
\centering
\small
\caption{Comparison of CT-1 training performance across different dataset settings (GS: General Scenarios; RS: Reasoning Scenarios). $a/b$ denotes an $a$ subset from the $b$ dataset.}
\resizebox{0.6\linewidth}{!}{
\begin{tabular}{l|l|cc|cc}
\toprule
  \multirow{2}{*}{Data} & \multirow{2}{*}{Scale (K)} &
  Imaging & Dynamic & Regular & Complex \\ 
  & & Quality & Degree & Speed & Motion  \\
  \midrule
             \textit{w/} GS  &  80 / 120  & \underline{0.674}  & \underline{0.782}   & \underline{80.9}  &  70.7    \\
             \textit{w/} RS  &  80 / 80  & 0.631  & 0.735    & 77.5  &  \underline{73.9}    \\
             \midrule
             CT-200K & 80 / 200 & \textbf{0.691} & \textbf{0.809} & \textbf{84.2} & \textbf{77.8} \\
\bottomrule
\end{tabular}}
\label{tab:data_efficiency}
\end{table}

\noindent \textbf{Data efficiency of CT-200K.}
Moreover, we conduct a systematic investigation into the training efficiency of general scenarios (denoted as GS) and Reasoning Scenarios (denoted as RS) in the CT-200K dataset. Specifically, we train the CT-1 model on different data subsets (all 80K) and perform a controlled comparative analysis. As reported in Table~\ref{tab:data_efficiency}, the inclusion of reasoning scenario data provides richer and more complementary supervisory signals for improving the model’s global understanding. As the proportion of this data increases, the model demonstrates a pronounced improvement in modeling complex camera motion.

\begin{table}[ht]
\centering
\small
\caption{Comparison of the time and memory costs of the CT-1 estimation stage and the video generation stage.}
\resizebox{0.6\linewidth}{!}{
\begin{tabular}{c|c|cc}
\toprule
\textbf{Model} & \textbf{Length} & \textbf{Time cost (seconds)} & \textbf{Memory (GB)} \\
\midrule
\rowcolor{lightgray!20}
CT-1 model & 13-step  & 22.81 & 28.91 \\
Video generation & 49-frame  & 429.02 & 39.48 \\
\bottomrule
\end{tabular}}
\label{tab:time_memory}
\end{table}

\noindent \textbf{Time and memory costs.}
We further analyze the time and memory costs of both the CT-1 estimation stage and the video generation stage on a single Nvidia device. Because the camera-controllable diffusion model adopts a 4× temporal compression, CT-1 predicts a 13-step camera trajectory for a 49-frame video. As reported in Table~\ref{tab:time_memory}, CT-1 introduces only a modest overhead relative to the downstream video generation stage, such as accounting for roughly 5.3\% of the total inference time.

%% file: section/conclusion.tex
\section{Conclusion}

We proposed a camera-controllable video generation framework within a new Vision–Language–Camera model termed CT-1. This model learns expressive camera motion distributions and transfers spatial reasoning knowledge into video generation for camera control. Supported by a large-scale curated dataset, CT-200K, our approach consistently produces high-quality camera-controllable videos, effectively bridging spatial reasoning and video synthesis.

%% file: section/appendix.tex
\section{Appendix}

\subsection{More Implementation Details}
\label{appendix:implementation_details}

\textbf{Evaluation metrics.}
The camera trajectories predicted by our proposed CT-1 model are inherently non-unique.
This means that multiple trajectories can satisfy the same semantic intent and physical constraints while differing in their numerical realizations. This property makes traditional point-wise error-based evaluation metrics inadequate. 
Therefore, we primarily adopt the ``success rate'' to evaluate the precision of camera control. This metric is widely used in the field of robotic control to measure the accuracy with which a system completes a specified task. In our setting, the success rate reflects whether the generated camera trajectories are semantically consistent with the given image–text conditions. Specifically, we construct an evaluation set of 100 image–text pairs, where the textual prompts cover a diverse range of camera motion types, \textit{e.g.,} ``the camera moves to the left,'' ``the camera zooms in,'' ``the camera rotates clockwise''.
After CT-1 generates the corresponding camera trajectories, two domain experts independently conduct manual evaluations of each generated trajectory.
A sample is counted as successful only if both experts unanimously agree that the trajectory correctly reflects the described camera motion. In cases where the two experts disagree, the sample is conservatively treated as a failure. The success rate is then computed as the ratio of successful samples to the total number of evaluated samples.

\noindent \textbf{WavReg loss.}
In WavReg loss, we use $[\lambda_a, \lambda_{d_L}, \lambda_{d_{L-1}}, \lambda_{d_1}]$ as the scale-dependent weights.
These values are set, for example, as $[\lambda_a, \lambda_{d_L}, \lambda_{d_{L-1}}, \lambda_{d_1}] = [2.0, 1.0, 0.5, 0.25]$.
It explicitly encourages the model to faithfully capture global camera motion trends and avoid overfitting to high-frequency noise. In extreme cases, we may set $\lambda_{d_l}=0$ and supervise only the low-frequency component, enforcing pure trend alignment.
Furthermore, in the WavReg calculation equation, \textit{i.e.,} the Eq. (8), the $\beta$ is a weight parameter (\textit{e.g.}, $\beta=0.1$) controlling the auxiliary supervision strength. 
Here, \textit{we summarize the properties of this design}: 1) The diffusion loss ensures correct noise modeling and generative diversity. 2) The wavelet loss injects structural inductive bias, guiding the model toward semantically stable camera motion. 3) Since the wavelet loss is applied, gradients propagate naturally through the noise prediction network without altering the sampling process.

\noindent \textbf{Experiment details.}
Our experimental details are divided into two parts as follows.

\noindent 1) \textit{Camera trajectory generation:}
Experiments were conducted using 32 NVIDIA GPUs. We trained the CT-1 model on the CT-200K dataset and evaluated its performance on CameraBench100~\cite{lin2025towards}, RealEstate10K~\cite{zhou2018stereo}, MultiCamVideo~\cite{bai2025recammaster}, and DrivingDoJo~\cite{wang2024drivingdojo} datasets. 
Each video sample is paired with a textual description specifying the associated camera motion.
In our experiments, for the CT-1 model, we extract the first frame of each video together with its corresponding textual description as the model input. For the video generation backbone, videos are synthesized conditioned on the textual descriptions and the extracted first frames. Within our framework, the same textual descriptions and initial frames are provided to the CT-1 model to estimate the corresponding camera trajectories.
All experiments were performed at a resolution of $224 \times 224$. The DiT blocks were trained end-to-end with a learning rate of $1 \times 10^{-5}$ and a batch size of 512.

\noindent 2) \textit{Integration of estimated camera trajectories into video diffusion models:}
Experiments were conducted on a single NVIDIA GPU for inference. We adopt CameraNoise~\cite{cameranoise}, CameraCtrl~\cite{he2025cameractrl}, and MotionCtrl~\cite{wang2024motionctrl} as the controllable video generation models, which take a reference image, a textual prompt, and the estimated camera trajectories as inputs.
Besides, in CameraBench100, most publicly available images are of relatively low resolution (\textit{e.g.,} $480 \times 270$), whereas current mainstream video generation backbone models are typically trained and evaluated at resolutions of 1024 or higher. To mitigate the impact of this resolution mismatch on performance evaluation, we reconstruct the reference images used as inputs to CT-1 and the video diffusion models. Specifically, based on the scene semantics and layout described in the original images, together with the corresponding textual prompts, we redraw the reference images using an existing open-source image generation model~\cite{cai2025z}. This procedure improves the input image quality and ensures that the experimental setting better reflects the intrinsic performance of high-resolution backbone models, thereby enabling a fair and reliable evaluation.

\subsection{More Qualitative Results}
\label{appendix:more_main_results}

\begin{figure*}[h]
    \centering
    \captionsetup{type=figure}
    \includegraphics[width=1.0\linewidth]{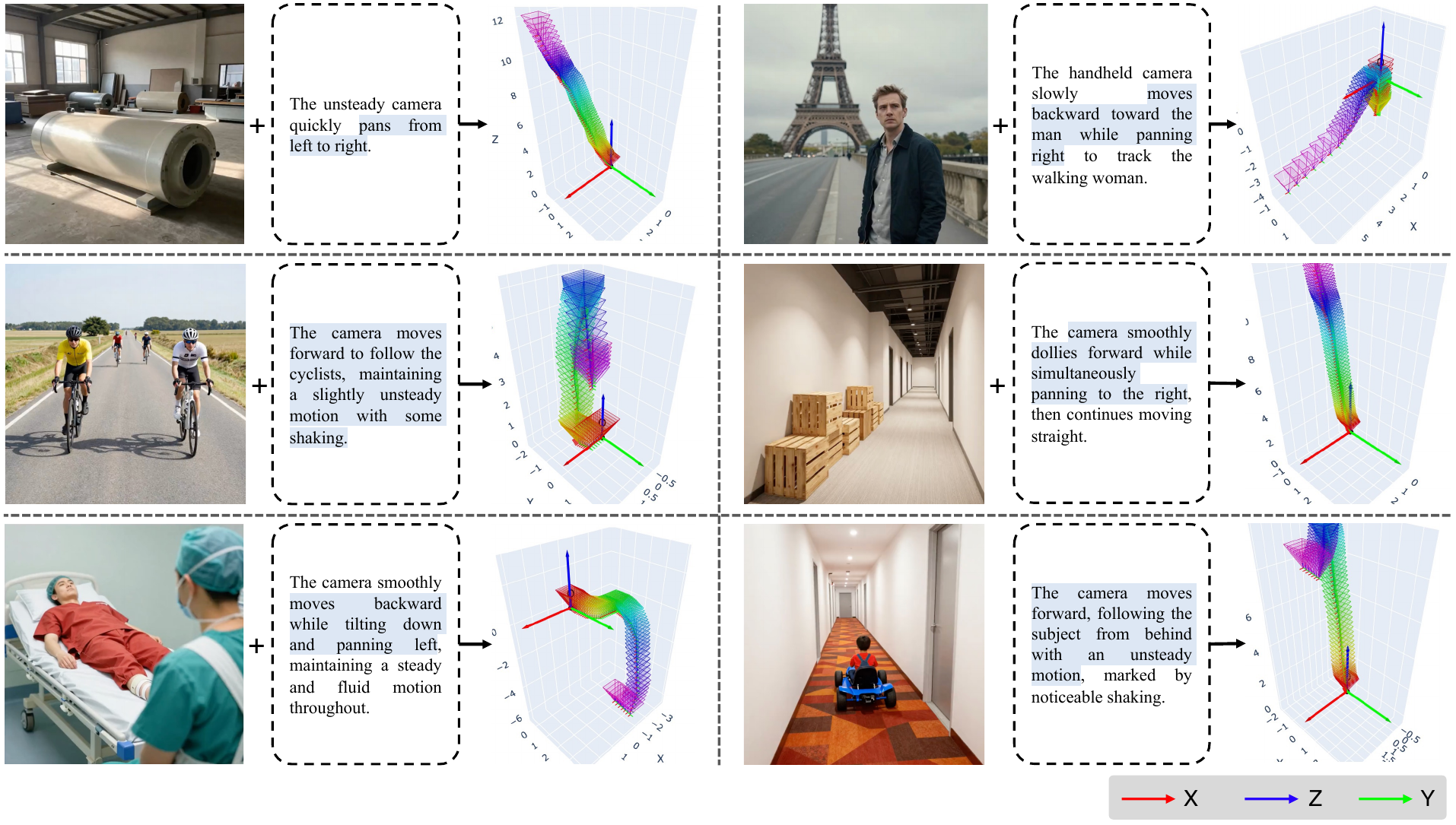}
    \caption{Visualization of camera trajectories estimated by CT-1. The red (X), green (Y), and blue (Z) axes indicate the camera’s local coordinate system~\cite{rockwell2025dynamic}: the red axis denotes the right–left direction, the green axis represents the vertical direction (down or up), and the blue axis corresponds to the viewing direction (forward or backward, \textit{i.e.,} zoom-in or zoom-out).}
    \label{fig:camera_trajectory_estimation_1}
\end{figure*}

\begin{figure*}[t!]
    \centering
    \captionsetup{type=figure}
    \includegraphics[width=1.0\linewidth]{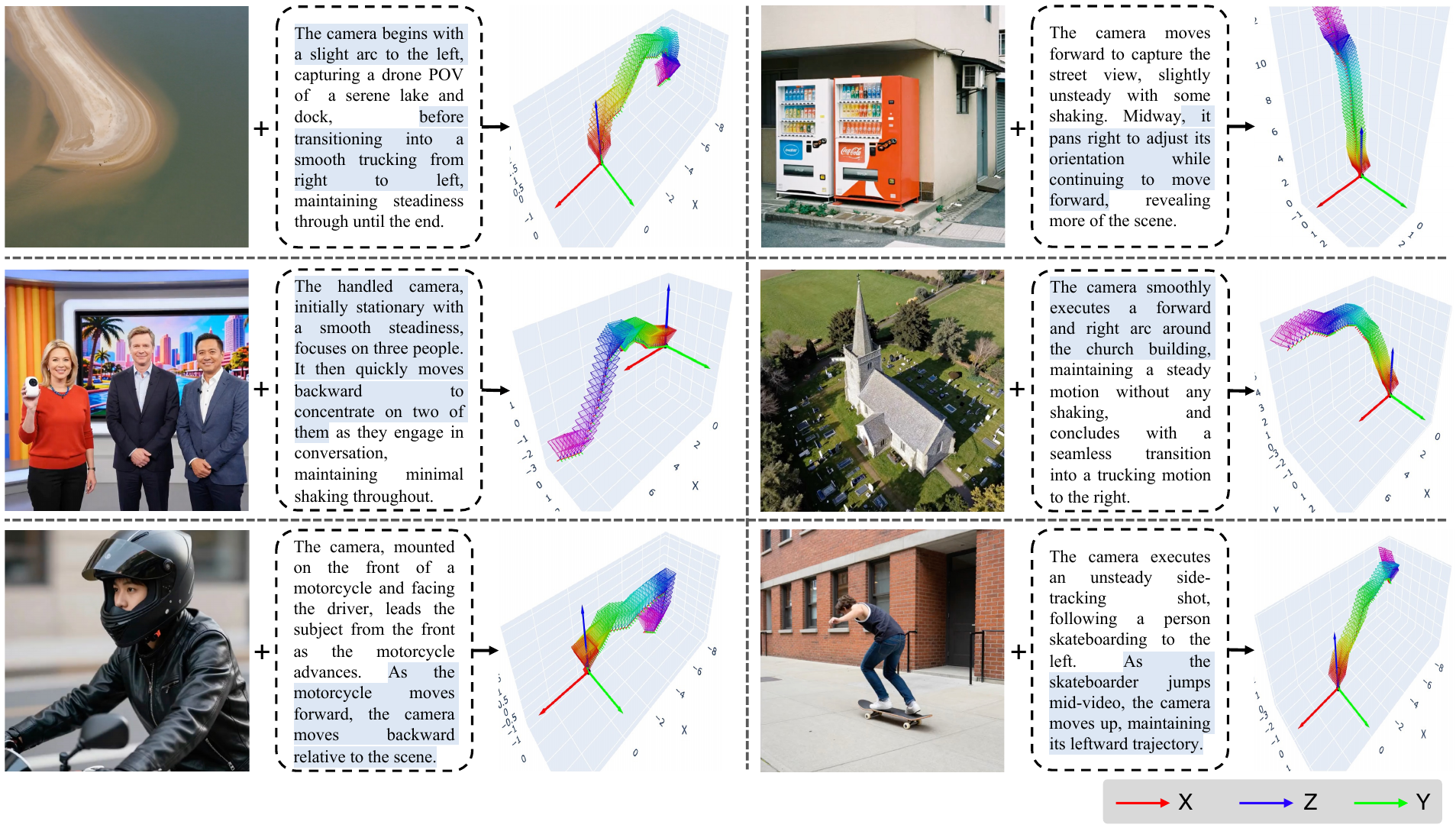}
    \caption{Visualization of camera trajectories estimated by CT-1 under longer textual descriptions.}
    \label{fig:camera_trajectory_estimation_2}
\end{figure*}

\begin{figure*}[t!]
    \centering
    \captionsetup{type=figure}
    \includegraphics[width=1.0\linewidth]{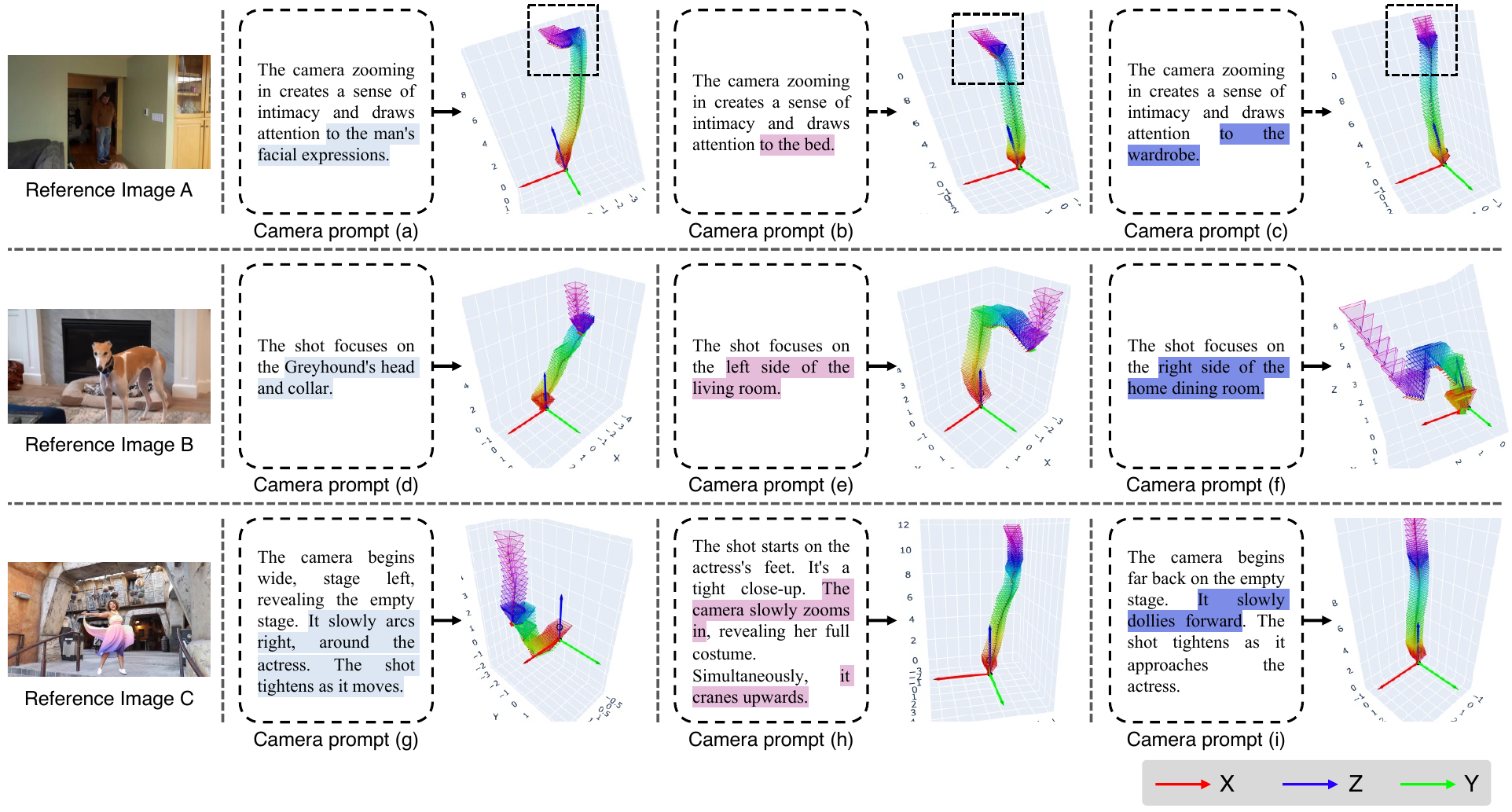}
    \caption{Visualization of a single image with multiple camera descriptions. Given a fixed reference image, we apply three different camera motion descriptions to the same scene and visualize the camera trajectories predicted by CT-1.}
    \label{fig:cross_camera_trajectory_estimation_1}
\end{figure*}

\subsubsection{Camera trajectories predicted by CT-1.}
\label{appendix:camera_trajectories}
When evaluating the visualization of the trajectories estimated by the CT-1 model, \textbf{it is important to note} that the trajectories inferred from the image and text are not unique. \textit{Therefore, the generated trajectories shown in this section are not compared with the ground truth.} As long as the trajectories accurately reflect the semantics of the image and text, they are considered valid. For ease of observation, we use the same coordinate system rules as in method~\cite{rockwell2025dynamic}, and \textbf{the physical meanings of these axes are:}
1) The red X-axis indicates the horizontal direction, where the arrow direction corresponds to rightward motion and the opposite direction corresponds to leftward motion. 2) The blue Z-axis aligns with the camera’s viewing direction, where the arrow direction represents forward (zoom-in) and the opposite direction represents backward (zoom-out). 3) The green Y-axis denotes the vertical direction, where the arrow direction corresponds to downward and the opposite corresponds to upward.

In Fig.~\ref{fig:camera_trajectory_estimation_1} and Fig.~\ref{fig:camera_trajectory_estimation_2}, we visualize the camera trajectory sequences estimated by CT-1 conditioned on the input reference image and textual description. Specifically, CT-1 first extracts the spatial structure of the scene from the input image and parses the semantic intent of camera motion from the text prompt. Under the joint conditioning of semantic understanding and spatial perception, the model predicts a temporally continuous sequence of camera poses, which characterizes the camera motion in three-dimensional space.
From the visualizations, we observe that the camera trajectories generated by CT-1 remain highly consistent with the input semantics in terms of global motion path, movement direction, and overall temporal evolution. 
In Fig.~\ref{fig:camera_trajectory_estimation_1}, which adopts standard-length textual prompts, CT-1 produces camera motions that closely follow the intended semantic guidance. 
In Fig.~\ref{fig:camera_trajectory_estimation_2}, we further evaluate CT-1 under more detailed and longer textual descriptions, aiming to assess its robustness to long-form semantic inputs. Notably, even under extended prompts, the predicted trajectories preserve strong semantic alignment while maintaining good temporal continuity and smoothness, without introducing abrupt changes or oscillatory artifacts. These results demonstrate that CT-1 can effectively translate high-level and long-range semantic constraints into structured camera motion representations, providing a stable and semantically aware camera trajectory prior for subsequent controllable video generation with diffusion models.

\begin{figure*}[t!]
    \centering
    \captionsetup{type=figure}
    \includegraphics[width=1.0\linewidth]{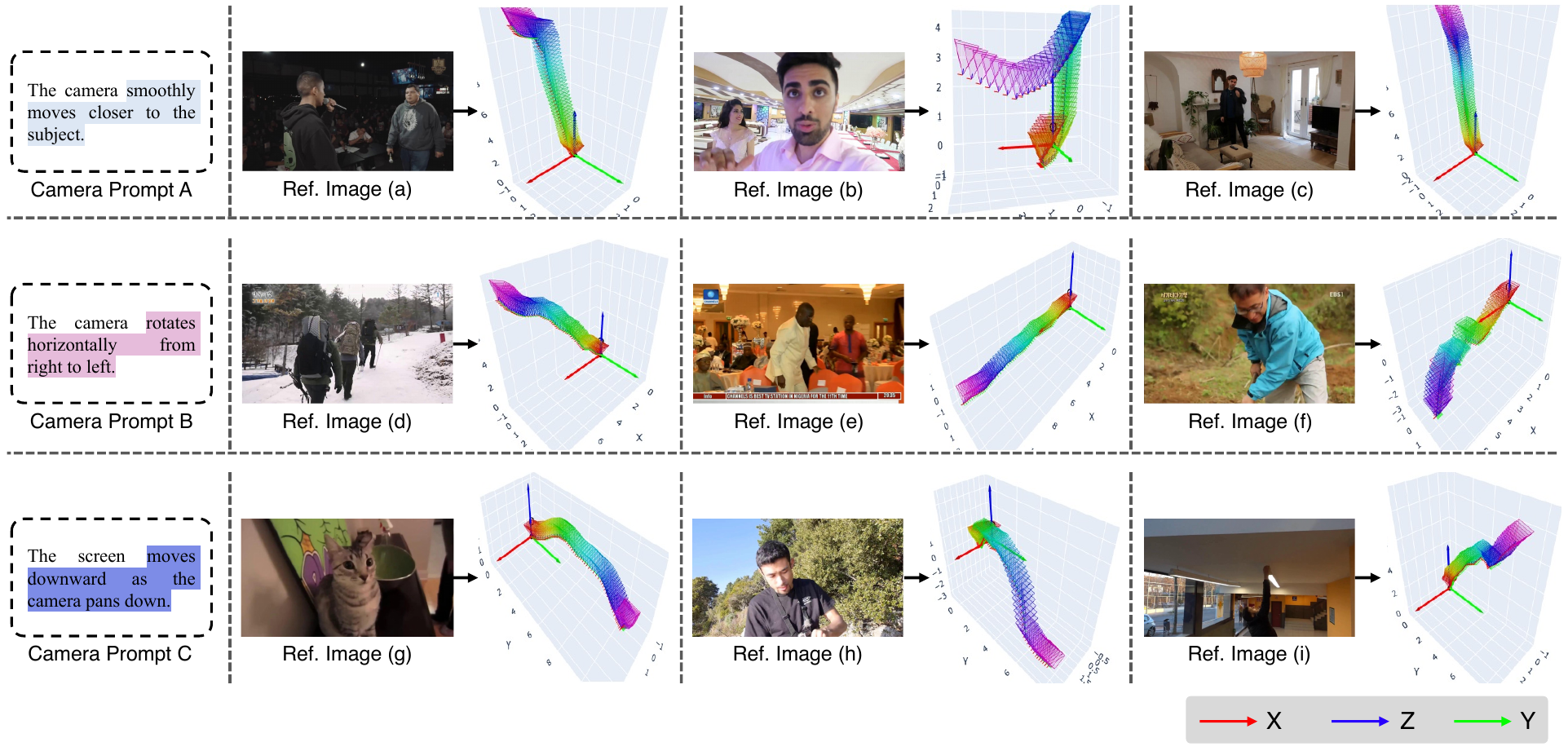}
    \caption{Visualization of a single camera description with multiple reference images. Given a fixed camera motion description, we pair it with three different reference images and infer the corresponding camera trajectories using CT-1.}
    \label{fig:cross_camera_trajectory_estimation_2}
\end{figure*}

\subsubsection{Multi-instance cross-validation of CT-1 with image and text inputs.}
\label{appendix:cross_validation}
In addition, to systematically evaluate the camera trajectory modeling capability of CT-1 under different image–text pairing conditions, we design two complementary cross-validation experiments that examine ``multiple camera descriptions with a single image'' and ``multiple images with a single camera description'', respectively.
Specifically, in the first setting, we fix a reference image $A$ and construct three camera motion descriptions with different semantic emphases and motion patterns to test the same scene. By comparing the predicted camera trajectories under different textual conditions (as shown in Fig.~\ref{fig:cross_camera_trajectory_estimation_1}), we analyze the sensitivity of CT-1 to variations in camera-related semantics and its ability to model diverse motion distributions given identical visual content.
In addition, in the experiments with reference image $A$, we further observe that even a minor modification of a key semantic word in the textual description (\textit{e.g.}, replacing ``man'' with ``bed'' in the same scene) can lead to a pronounced change in the predicted camera parameters. In this case, the camera first moves forward and then turns to focus on different target objects, exhibiting clearly different viewpoint transition behaviors under different textual conditions. This observation indicates that CT-1 is highly sensitive to fine-grained semantic variations in the text prompts, and is able to effectively translate subtle linguistic differences into distinguishable camera motion decisions, demonstrating strong responsiveness and semantic controllability in language-guided camera control.

In the second setting, we fix a single camera description $\hat{A}$ and pair it with three reference images exhibiting different scene layouts and geometric structures. These image–text pairs are independently fed into CT-1 to infer camera trajectories, and the results are reported in Fig.~\ref{fig:cross_camera_trajectory_estimation_2}. This setting is used to evaluate the adaptability of CT-1 to different visual contexts under the same semantic constraint, as well as its capability to maintain consistent motion semantics across heterogeneous scenes.
Together, these two experimental settings provide complementary perspectives to comprehensively assess the robustness, interpretability, and semantic alignment of CT-1 when both visual and linguistic conditions vary.

\begin{figure*}[t!]
    \centering
    \captionsetup{type=figure}
    \includegraphics[width=1.0\linewidth]{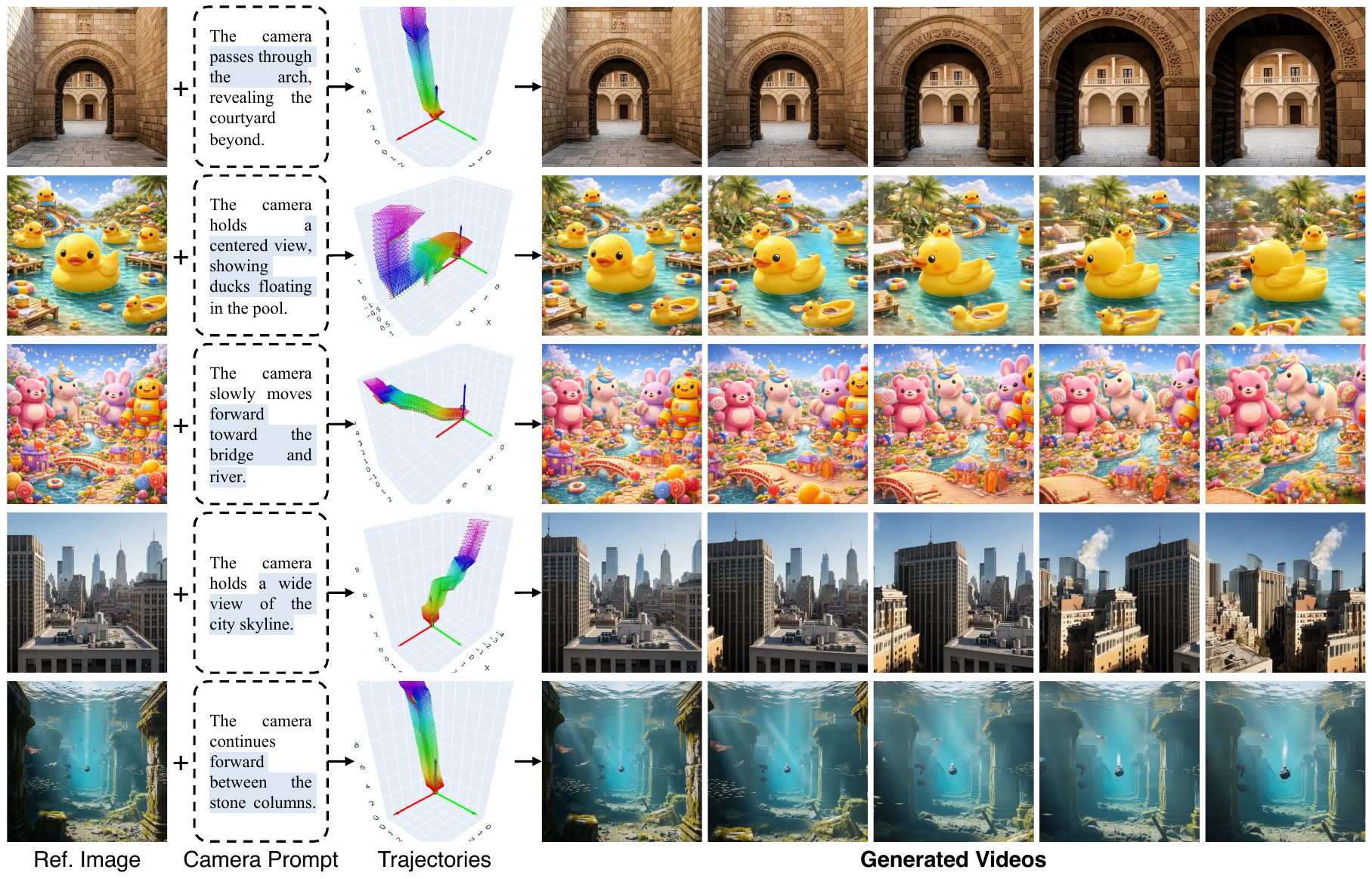}
    \caption{Qualitative results of generated videos. Given an input reference image and a textual prompt, CT-1 infers semantically consistent camera motion trajectories, which are further used to drive a camera-controllable video generation model~\cite{cameranoise}. In each case, scene-related elements are highlighted with a light blue background to indicate the visual content explicitly referenced by the input camera description within the user intent.}
    \label{fig:challenging_case}
\end{figure*}

\subsubsection{Performances on challenging scenarios.}
\label{appendix:challenging_scenarios}
To systematically evaluate the reasoning capability of CT-1 under complex scenarios and high-level semantic conditions, we test CT-1 on five representative scenes in Fig.~\ref{fig:challenging_case}, and further generate the corresponding videos based on the predicted camera trajectories. These scenes cover diverse and challenging visual contents and semantic structures, including non-photorealistic settings with cartoon characters, which impose higher requirements on cross-style generalization and semantic understanding.
From the qualitative results, we observe that even in such complex and abstract scenarios, CT-1 can accurately infer camera motion trajectories conditioned on the input image and textual prompt. The predicted trajectories remain highly consistent with the user intent in terms of global path, motion direction, and overall evolution trend. Moreover, when integrating these trajectories with a camera-controllable video generation model, the resulting videos successfully exhibit camera motions that faithfully follow the predicted trajectories, demonstrating the effectiveness and stability of CT-1 in driving downstream video generation under complex conditions.

\begin{figure*}[t!]
    \centering
    \captionsetup{type=figure}
    \includegraphics[width=1.0\linewidth]{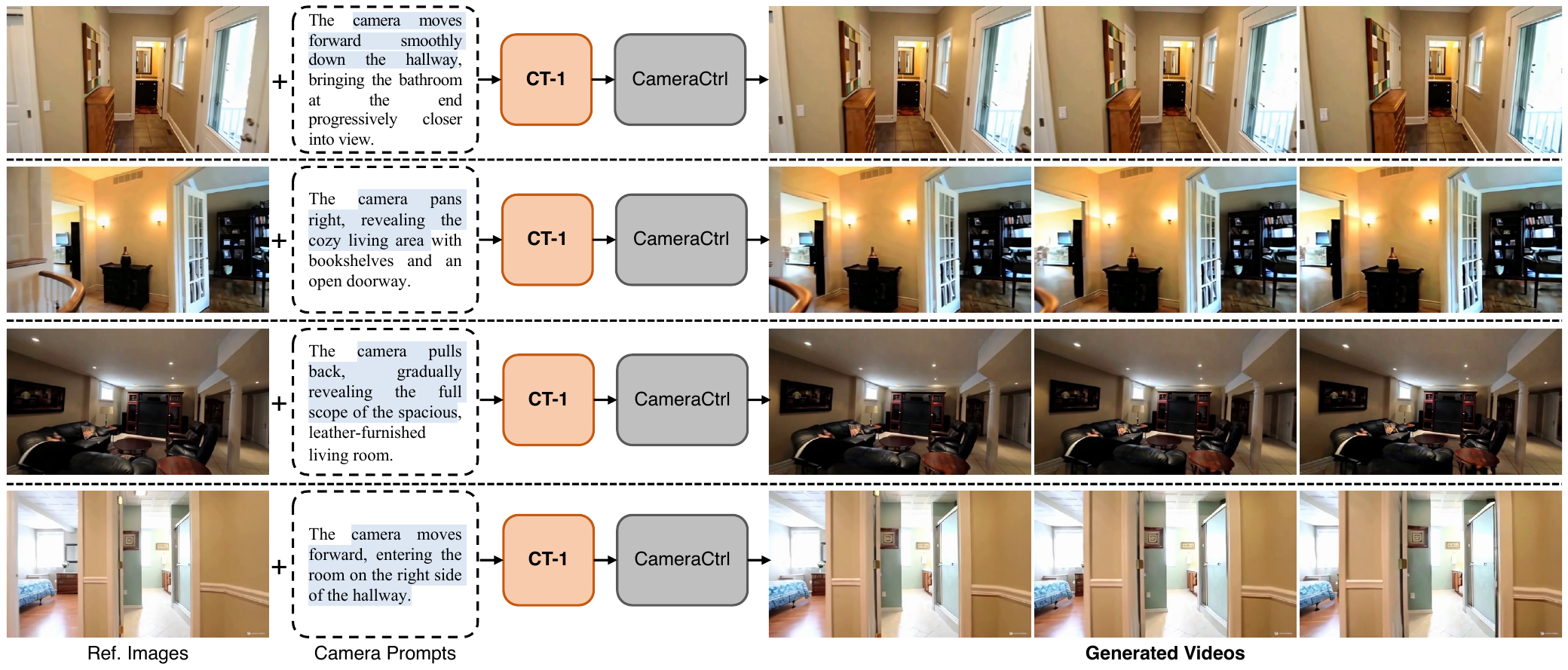}
    \caption{Qualitative results of applying the trajectories results of the CT-1 on CameraCtrl~\cite{he2025cameractrl} model, which is a camera-controllable video generation model. The tested scenes are from the RealEstate10K dataset~\cite{zhou2018stereo}.}
    \label{fig:Apple_to_other_methods_1}
\end{figure*}

\begin{figure*}[t!]
    \centering
    \captionsetup{type=figure}
    \includegraphics[width=1.0\linewidth]{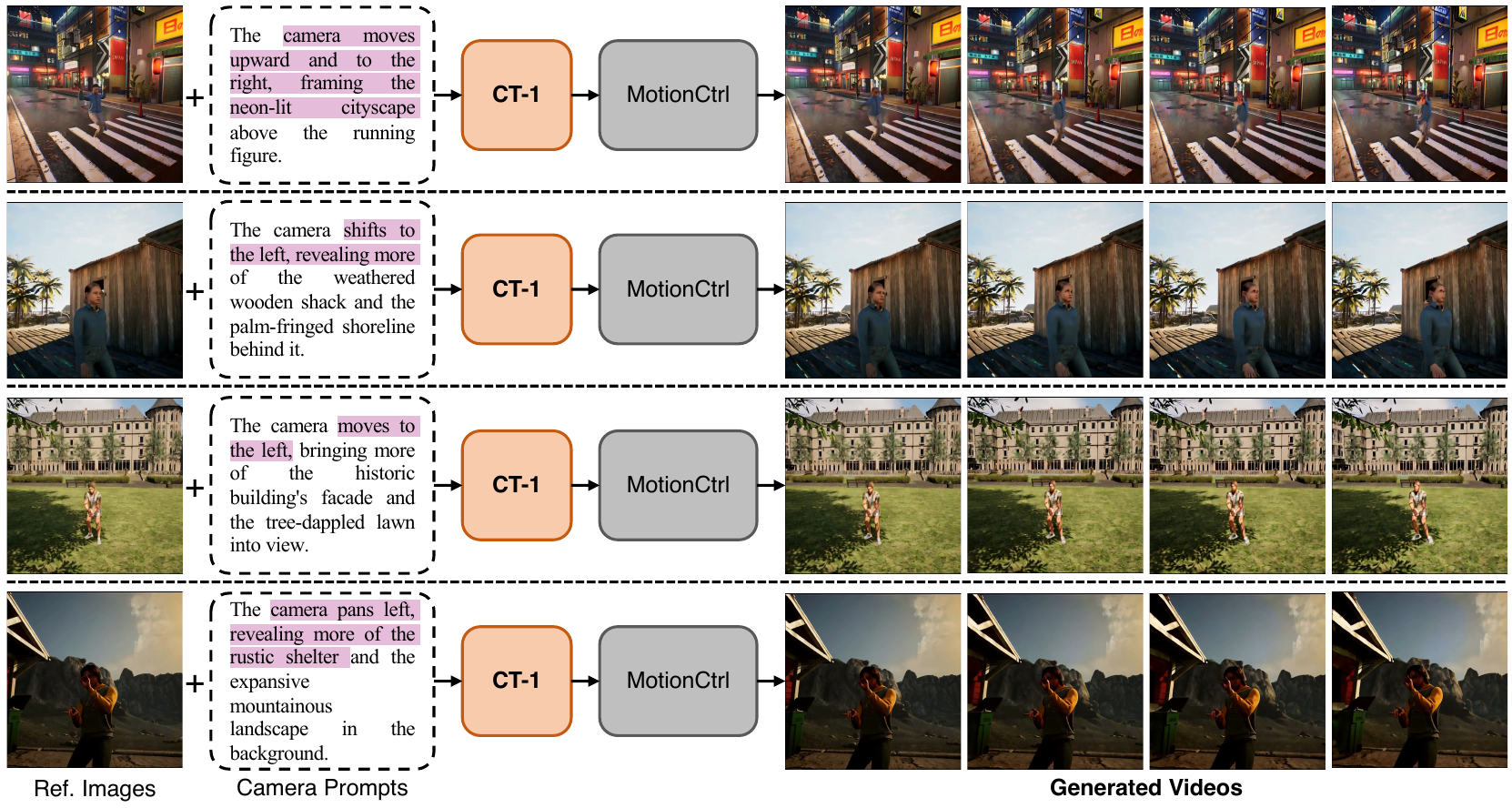}
    \caption{Qualitative results of applying the trajectories results of the CT-1 on MotionCtrl~\cite{wang2024motionctrl} model, which is a camera-controllable video generation model. The tested scenes are from the MultiCamVideo dataset~\cite{bai2025recammaster}.}
    \label{fig:Apple_to_other_methods_2}
\end{figure*}

\subsubsection{Applying CT-1 to other camera-controllable video diffusion models.}
\label{appendix:applying2others}
In this paper, we present a novel camera-controllable video generation framework aimed at significantly enhancing the control of camera motion, particularly in terms of spatial-semantic understanding during the video generation process. This capability is successfully transferred into the video generation framework through the introduction of the CT-1 model, enabling the generated videos to more accurately follow predefined camera motion trajectories. 
Besides the extensive experiments within the CameraNoise framework~\cite{cameranoise}, we also integrated the CT-1 model with two other methods, CameraCtrl~\cite{he2025cameractrl} and MotionCtrl~\cite{wang2024motionctrl}, further enhancing the accuracy and consistency of the generated results.

Specifically, the CT-1 model estimates high-quality camera trajectories based on an in-depth understanding of scene information and textual semantic cues, which are then used as control signals to drive the CameraCtrl and MotionCtrl methods for camera-controllable video generation.
Fig.~\ref{fig:Apple_to_other_methods_1} illustrates the performance of the CameraCtrl on the RealEstate10K dataset~\cite{zhou2018stereo}.
Furthermore, Fig.~\ref{fig:Apple_to_other_methods_2} demonstrates the effectiveness of the MotionCtrl method on the MultiCamVideo dataset~\cite{bai2025recammaster}, further validating the successful combination of CT-1 with these two methods. 
So, the proposed CT-1 model not only has independent advantages but can also be integrated with existing methods, thereby optimizing and guiding current camera-controllable video generation models.
Moreover, due to inherent differences in camera control capability across models, we observe that MotionCtrl exhibits a smaller motion magnitude than CameraCtrl.

\begin{figure*}[t!]
    \centering
    \captionsetup{type=figure}
    \includegraphics[width=1.0\linewidth]{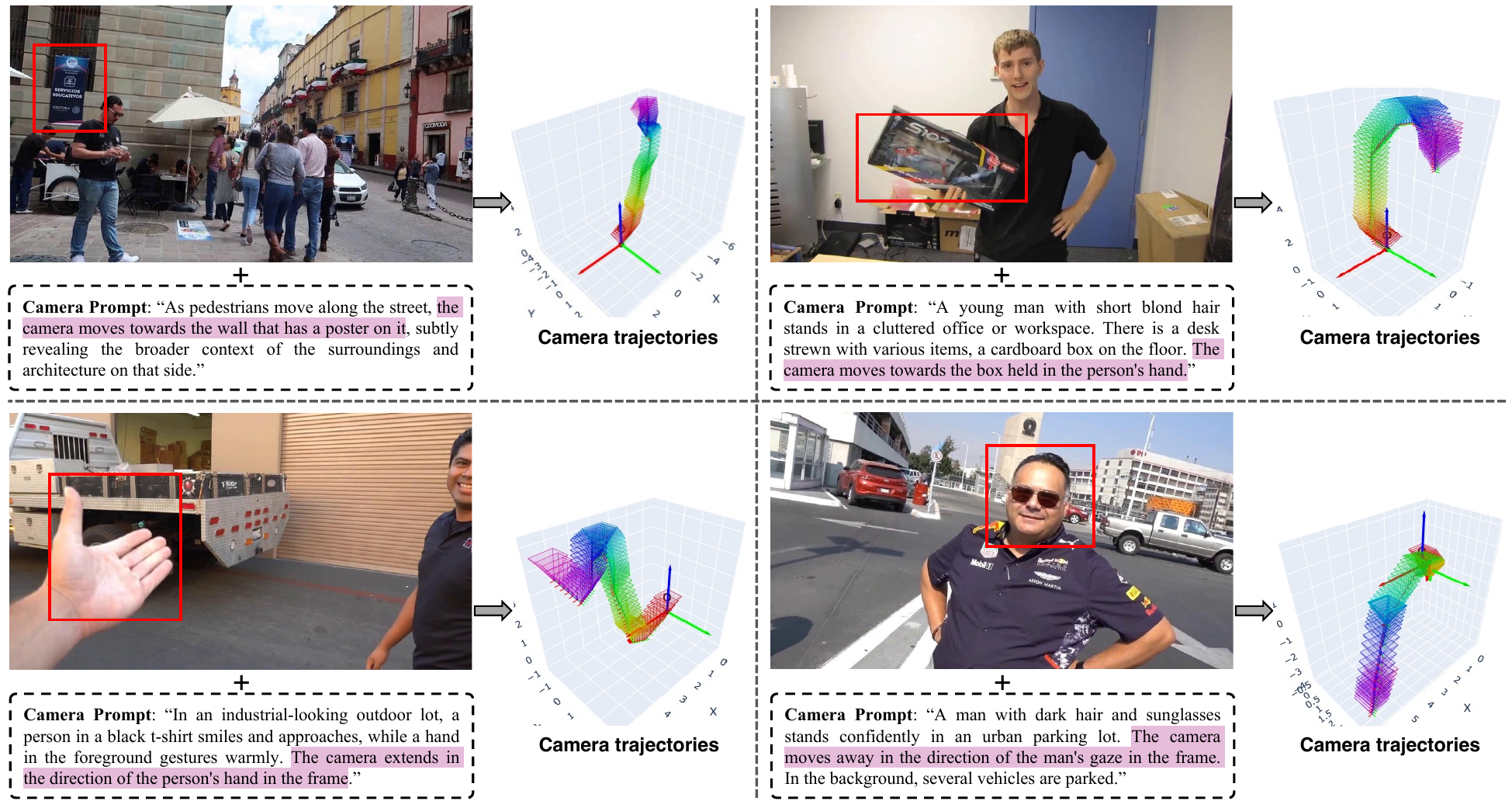}
    \caption{Qualitative results of deep reasoning-based setting, showcasing the CT-1 model’s ability to infer camera motion from semantic and visual clues in complex scenes. The red boxes highlight implicit cues in the text that require visual context for interpretation.}
    \label{fig:Reasoning_case}
\end{figure*}

\begin{figure*}[t!]
    \centering
    \captionsetup{type=figure}
    \includegraphics[width=1.0\linewidth]{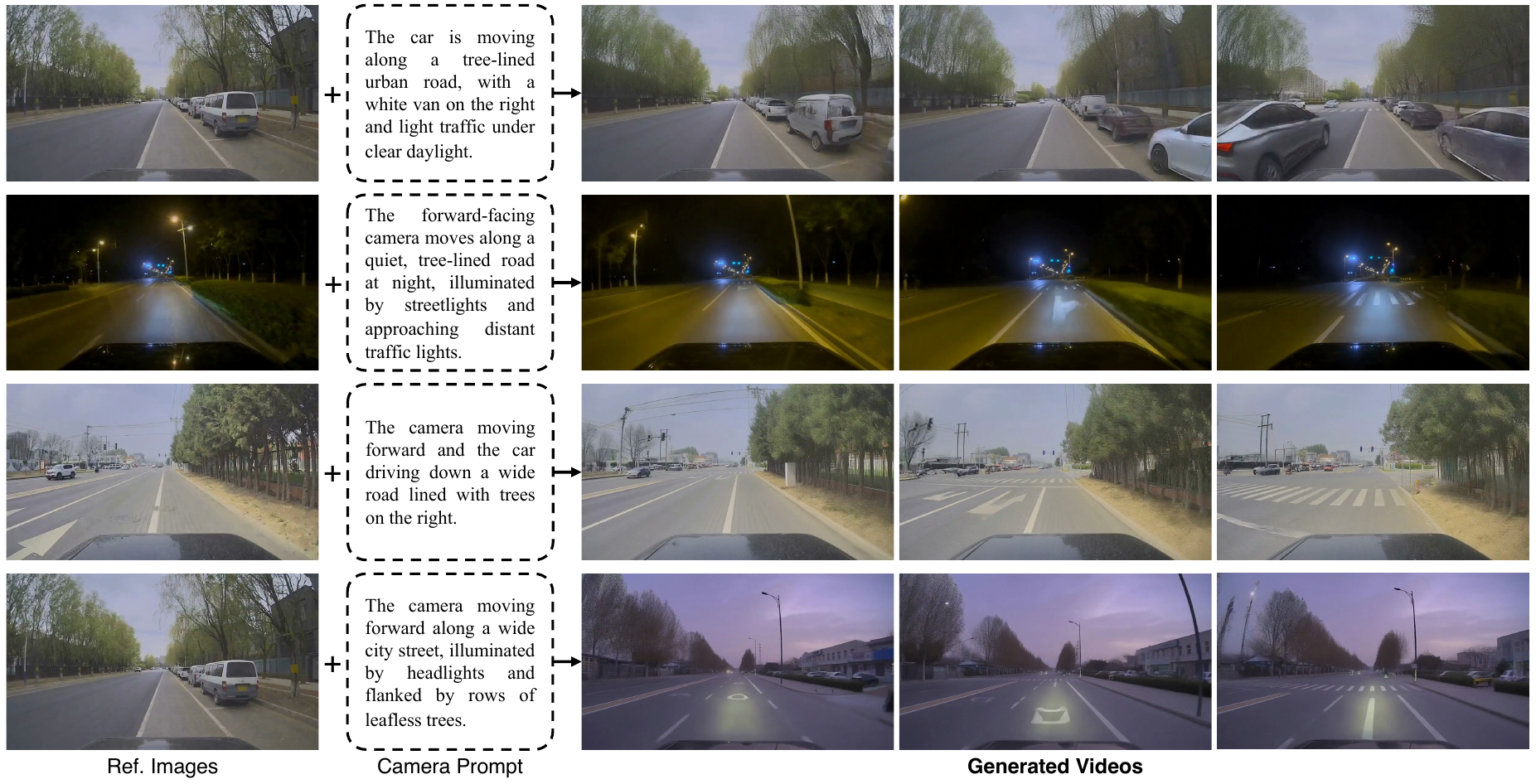}
    \caption{Qualitative results of camera-controllable video generation in driving scenarios on DrivingDoJo dataset~\cite{wang2024drivingdojo}. For each example, we provide a reference image and a forward-driving textual prompt as input to CT-1, which predicts the corresponding camera trajectory and drives the subsequent video diffusion model for synthesis.}
    \label{fig:driving_scene}
\end{figure*}

\subsubsection{Deep-reasoning-based camera trajectory generation.}
\label{appendix:deep_reasoning}
In addition, to validate the model's reasoning ability in combining visual content and semantics, we design deep reasoning-based experiments. As shown in Fig.~\ref{fig:Reasoning_case}, in scenes with rich visual elements, we add descriptive text to guide the reasoning, rather than explicitly specifying whether the camera should move left or right. Instead, the model infers the motion based on clues within the scene. For instance, in the example in the top-left corner, the camera is described as moving toward a wall with posters. This requires the model to understand which areas of the scene contain walls and to locate the wall with the poster. From other examples, it is evident that the proposed CT-1 model demonstrates strong visual reasoning abilities, being able to infer how the camera should move based on the clues in the scene.

\subsubsection{Application extension: driving scenarios.}
\label{appendix:driving_scenarios}
As a further extension of camera-controllable video generation, we apply the proposed CT-1 model to real-world driving scenarios to jointly estimate camera parameters and synthesize corresponding videos. The qualitative results are presented in Fig.~\ref{fig:driving_scene}. Specifically, we collect representative driving samples from the DrivingDoJo dataset~\cite{wang2024drivingdojo} and design targeted textual descriptions centered on the canonical ``forward-driving'' camera pattern, by taking into account the road layout, motion direction, and surrounding environmental structure in driving scenes.
Under this setting, CT-1 is able to infer physically plausible and semantically consistent camera trajectories from a given reference image and textual prompt, and subsequently transfer the predicted trajectories to a camera-controllable video diffusion model for video synthesis. As shown in the figure, even in driving environments with strong geometric constraints and complex scene structures, our method can robustly perform camera parameter estimation and video generation, accurately capturing the global forward motion and viewpoint evolution along the driving direction.
These results demonstrate that CT-1 is not limited to generic or reasoning scenarios, but also exhibits strong cross-scenario generalization capability, enabling effective camera control in realistic driving applications with clear motion patterns and strict temporal constraints.

\begin{figure}[t!]
    \centering
    \captionsetup{type=figure}
    \includegraphics[width=0.6\linewidth]{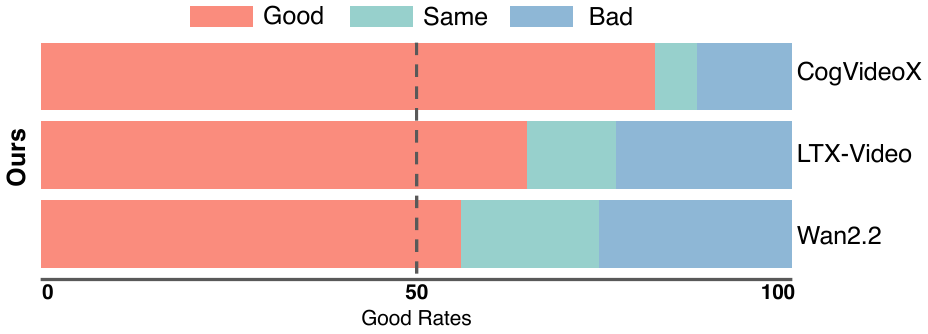}
    \caption{User preference on 30 selected videos. Longer pink bars indicate greater satisfaction with our method.}
    \label{fig:user_study}
\end{figure}

\subsubsection{User study.}
\label{appendix:user_study}
We conduct a user preference study against CogVideoX~\cite{yang2024cogvideox}, LTX-Video~\cite{hacohen2024ltx}, and Wan2.2~\cite{wan2025wan}, as shown in Fig.~\ref{fig:user_study}. Human evaluators are asked to compare videos generated by different methods in a pairwise manner. We evaluate the results from four aspects: per-frame image quality, camera control accuracy, motion fidelity, and overall video quality. As shown in Fig.~\ref{fig:user_study}, our method achieves the best preference scores across all criteria, indicating superior visual quality, motion consistency, and camera controllability.

\subsection{More Explanation of VLC Model}
\label{appendix:vlc_explanation}

\subsubsection{Discussion of the VLC model.}
The VLC model predicts scene-aware camera trajectories from image-text conditions for controllable video generation. Its goal is to model camera motion under the joint constraint of visual context and language intent, rather than to perform a generic multimodal regression. In camera control, the same textual instruction may correspond to different valid trajectories depending on scene layout, viewpoint, object arrangement, and spatial depth cues in the reference image. The model therefore jointly encodes the reference image and the user instruction, and generates a temporally coherent trajectory that is semantically aligned with the desired motion. The predicted trajectory is further used as an explicit intermediate control representation for downstream video synthesis, which enables camera motion to be modeled separately from the generation process and supports more precise control.

Compared with existing directions such as text-to-camera motion~\cite{lin2025towards}, camera trajectory estimation~\cite{li2023two,yu2025trajectorycrafter,wang2024tram}, and generic vision-language regression~\cite{wei2024vary,kim2025fine}, the proposed VLC model addresses a more specific generation-oriented setting:

\begin{itemize}[leftmargin=*, noitemsep, topsep=10pt]
    \item Text-to-camera motion methods mainly infer camera movement from language alone, while VLC jointly conditions the trajectory on both the reference image and the textual instruction, since the same camera command may correspond to different valid motions under different scene layouts, viewpoints, and spatial configurations.
    
    \item Camera trajectory estimation focuses on recovering camera motion from observed videos or image sequences, whereas VLC predicts a plausible future trajectory before generation and uses it as an explicit control signal for synthesis.
    
    \item Generic vision-language regression typically maps multimodal inputs to unconstrained continuous targets, while VLC models a temporally coherent camera trajectory whose output must satisfy semantic alignment, motion smoothness, and controllability requirements for downstream video generation.
\end{itemize}

Therefore, the VLC model introduces an explicit camera modeling framework that connects multimodal understanding with controllable video synthesis through a structured intermediate representation.

\subsubsection{Camera properties under the VLC model.}
\label{appendix:camera_properties}
In the main manuscript, we emphasize that the camera trajectories estimated by the VLC model exhibit the following properties:

\noindent 1) \textit{Semantic Alignment}, where the trajectories conform to the motion semantics of the vision–language instructions and form continuous curves on the $SE(3)$.

\noindent 2) \textit{Temporal Continuity}, where the trajectories are Lipschitz continuous:

\begin{equation}
d_{SE(3)}(K_t,K_{t-1})\leq L\cdot\Delta t,\quad\forall t
\end{equation}

Here, $d_{SE(3)}$ denotes the geodesic distance on $SE(3)$, $L$ is the Lipschitz constant, and $\Delta t$ represents the inter-frame time step. This property ensures smooth and stable camera motion.

\noindent 3) \textit{non-uniqueness}, where multiple semantically valid trajectories may exist for the same instruction:

\begin{equation}
\mathrm{Var}_{p_\theta}(K_{1:T}\mid v_{1:T},\ell)\mathrm{~}>\mathrm{~}0.
\end{equation}

Therefore, we define such trajectories as a distribution rather than a single curve, reflecting the inherent diversity of motion. With our proposed Diffusion Transformer-based CT-1 model, we can achieve flexible and diverse trajectory generation.

\subsection{More Ablation Studies}

\begin{table}[h]
\centering
\small
\caption{Comparison of camera trajectory quality with different CT-1 variants. The second-best results are underlined.}
\resizebox{0.5\linewidth}{!}{
\begin{tabular}{l|cc|cc}
\toprule
  \multirow{2}{*}{Method} &
  Aesthetic & Imaging & Regular & Complex \\ 
  & Quality & Quality & Speed & Motion \\
  \midrule
             $\textrm{CT-1}^{*}$-GR  & 0.557  & 0.681    & 83.9     &  76.8  \\ 
             $\textrm{CT-1}^{*}$-AR  & \underline{0.573}  & \underline{0.688}    & \underline{85.1}     &  \underline{79.4}   \\
             \midrule
             \textbf{CT-1}  & \textbf{0.585}  & \textbf{0.709}  & \textbf{87.6}  & \textbf{81.5} \\
\bottomrule
\end{tabular}}
\label{tab:diffusion_transformer}
\end{table}

\subsubsection{Effect of diffusion transformer.}
To justify the use of a diffusion transformer for trajectory modeling, we compare CT-1 with several simpler alternatives under the same image-text input and training protocol. Specifically, we replace the diffusion trajectory model with two other variants:

\begin{itemize}[leftmargin=*, noitemsep, topsep=10pt]
    \item $\textrm{CT-1}^{*}$-GR: a direct Gaussian Regression head that predicts a deterministic trajectory, and
    \item $\textrm{CT-1}^{*}$-AR: an Autoregressive transformer that predicts trajectory tokens step by step.
\end{itemize}

All variants use the same vision-language encoder and the same camera-context token design whenever applicable, so that the comparison isolates the effect of the trajectory generation paradigm.
We evaluate these variants on both trajectory prediction and downstream camera-controllable video generation. Since camera trajectories are inherently non-unique, we primarily report success rate on CameraBench100, together with the quality of generated videos.
Table~\ref{tab:diffusion_transformer} shows that the diffusion transformer is particularly beneficial because camera motion estimation is a one-to-many problem: multiple trajectories can satisfy the same semantic instruction while differing in numerical details. Deterministic regression tends to average over plausible motions, and autoregressive prediction is more prone to error accumulation over long horizons. In contrast, the diffusion transformer models the full trajectory distribution in a holistic manner, leading to better controllability and temporal stability.

\begin{table*}[ht]
\centering
\small
\caption{Statistical stability analysis of CameraBench100. We compare CT-1 and Wan2.2 across different test sets.}
\resizebox{1.0\linewidth}{!}{
\begin{tabular}{l c l c c c c}
\toprule
\textbf{Evaluation Setting} & \textbf{Size} & \textbf{Protocol} & \textbf{CT-1} & \textbf{Wan2.2~\cite{wan2025wan} w/ PE} & \textbf{Gap} & \textbf{Stability} \\
\midrule
\rowcolor{lightgray!20}
Fixed CameraBench100
& 100
& Fixed balanced subset
& 81.6
& 64.9
& +16.7
& -- \\
\midrule

Bootstrap CI
& 100
& 1000 bootstrap resamples
& 81.6 [74.8, 87.9]
& 64.9 [56.8, 72.7]
& +16.7
& CI reported \\
\midrule

Random balanced subsets
& 100
& Mean over 20 subsets
& 80.9 $\pm$ 1.7
& 63.8 $\pm$ 2.2
& +17.1 $\pm$ 1.1
& CT-1 wins 20/20 \\
\midrule

Larger balanced subsets
& 200
& Mean over 10 subsets
& 81.2 $\pm$ 1.1
& 64.3 $\pm$ 1.5
& +16.9 $\pm$ 0.8
& CT-1 wins 10/10 \\

Larger balanced subsets
& 300
& Mean over 10 subsets
& 81.0 $\pm$ 0.9
& 64.5 $\pm$ 1.2
& +16.5 $\pm$ 0.6
& CT-1 wins 10/10 \\
\bottomrule
\end{tabular}}
\label{tab:benchmark_analysis}
\end{table*}

\subsubsection{Statistical stability of CameraBench100.}
To comprehensively analyze the statistical stability and potential selection bias, we conduct additional analyses to examine whether the advantage of CT-1 remains consistent under different evaluation protocols. Specifically, we first clarify that CameraBench100 is a fixed balanced subset constructed to cover diverse camera motion types and scene categories.
Based on this benchmark, we perform four complementary analyses: 
1) we report the bootstrap confidence interval (Bootstrap CI) of the success rate on the fixed CameraBench100 subset, where the interval is estimated by repeated bootstrap resampling.
2) we repeat the evaluation on multiple randomly sampled balanced subsets of the same size to assess robustness to subset selection.
3) we further evaluate on larger balanced subsets when available, in order to examine whether the observed advantage persists as the evaluation size increases.
4) we report the performance gap (Gap), defined as the difference in success rate between CT-1 and the strongest baseline, together with a consistency statistic indicating whether CT-1 consistently outperforms the baseline across repeated subset samplings.

As shown in Table~\ref{tab:benchmark_analysis}, CT-1 achieves 81.6 average success rate on the fixed subset, outperforming ``Wan2.2 w/ PE'' by 16.7 points. This advantage remains stable under bootstrap resampling, where CT-1 consistently maintains a clear margin over the Wan2.2. 
``CT-1 wins 20/20'' denotes that CT-1 wins on all 20 / 20 random subsets, outperforming the Wan2.2 in every trial.
We further repeat the evaluation on multiple balanced random subsets of the same size and on larger balanced subsets. Across all these settings, CT-1 shows highly consistent performance and outperforms the baseline on every sampled subset. These results suggest that the superiority of CT-1 is statistically stable, robust to subset selection, and not tied to a particular choice of evaluation examples.

\begin{table}[t!]
\centering
\small
\caption{Comparison of different trajectory regularization strategies. The second-best results are underlined.}
\resizebox{0.5\linewidth}{!}{
\begin{tabular}{l|cc|cc}
\toprule
  \multirow{2}{*}{Method} &
  Aesthetic & Imaging & Regular & Complex \\ 
  & Quality & Quality & Speed & Motion \\
  \midrule
             $\textrm{VelReg}$  & 0.559  & 0.683    & 84.1     &  76.9  \\
             $\textrm{AccReg}$  & 0.562  & \underline{0.705}    & 85.7     &  79.1  \\
             $\textrm{Jerk}$    & 0.567  & 0.702    & 86.5     &  80.0  \\
             $\textrm{LowPass}$  & \underline{0.570}  &  0.696   & \underline{87.0}     &  \underline{80.7}  \\
             \midrule
             \textbf{$\textrm{WavReg}$}  & \textbf{0.585}  & \textbf{0.709}  & \textbf{87.6}  & \textbf{81.5} \\
\bottomrule
\end{tabular}}
\label{tab:trajectory_regularization}
\end{table}

\subsubsection{Effect of frequency-aware modeling.}
We further investigate whether the benefit of WavReg truly arises from its wavelet-based frequency-domain regularization, rather than from simply imposing a generic smoothing prior on the trajectory. To this end, on CameraBench100, we compare WavReg with several standard temporal smoothness objectives, including velocity regularization ($\textrm{VelReg}$), acceleration regularization ($\textrm{AccReg}$), and jerk penalty ($\textrm{Jerk}$), which are defined as the squared $L2$ norms of the first-, second-, and third-order temporal differences of the predicted trajectory, respectively. We also include a low-pass regularization baseline ($\textrm{LowPass}$), where the predicted trajectory is transformed into the frequency domain via the Fourier transform and the energy of high-frequency components beyond a fixed cutoff is penalized, without using the proposed wavelet formulation. For a fair comparison, all these baselines are incorporated into the same training objective as WavReg by replacing only the proposed regularization term, while keeping the backbone, trajectory prediction loss, and training protocol unchanged.

Shown in Table~\ref{tab:trajectory_regularization}, the results show that WavReg is not equivalent to a simple smoothing prior. Compared with velocity or acceleration penalties, WavReg better preserves the intended low-frequency motion trend while suppressing undesirable local oscillations, resulting in smoother yet still responsive trajectories. In contrast, simple smoothness losses tend to over-penalize legitimate motion changes and may reduce control accuracy.

\subsubsection{Effect of camera-context token.}
We ablate the contribution of the proposed camera-context token by varying both the input modalities and the conditioning mechanism. Specifically, we compare the proposed $\textrm{<CAM> Modeling}$ with:

\begin{itemize}[leftmargin=*, noitemsep, topsep=10pt]
    \item $\textrm{Text-only Modeling}$: text-only conditioning,
    \item $\textrm{Vision-only Modeling}$: image-only conditioning, and
    \item $\textrm{Pooled Modeling}$: image-text conditioning without the dedicated <CAM> token, where the pooled visual-language feature is directly injected into the trajectory model.
\end{itemize}

\begin{table}[t!]
\centering
\small
\caption{Comparison of different camera-context token strategies. The second-best results are underlined.}
\resizebox{0.6\linewidth}{!}{
\begin{tabular}{l|cc|cc}
\toprule
  \multirow{2}{*}{Method} &
  Aesthetic & Imaging & Regular & Complex \\ 
  & Quality & Quality & Speed & Motion \\
  \midrule
             $\textrm{Text-only Modeling}$  & 0.453  & 0.578    & 41.7     &  33.9  \\
             $\textrm{Vision-only Modeling}$  & 0.546  & 0.671    & 40.4     &  32.1  \\
             $\textrm{Pooled Modeling}$    & \underline{0.579}  & \underline{0.683}    & \underline{82.8}     &  \underline{76.2}  \\
             \midrule
             \textbf{$\textrm{<CAM> Modeling}$}  & \textbf{0.585}  & \textbf{0.709}  & \textbf{87.6}  & \textbf{81.5} \\
\bottomrule
\end{tabular}}
\label{tab:camera_context_token}
\end{table}

This ablation aims to answer two questions. First, how much of the performance comes from jointly using visual and linguistic cues? Second, is the dedicated camera-context token necessary, or can generic pooled conditioning achieve a similar effect? The comparison is conducted under the same backbone and tested on CameraBench100.
From Table~\ref{tab:camera_context_token}, we find that the proposed $\textrm{<CAM> Modeling}$ performs best.
These results demonstrate that text-only conditioning can capture coarse motion intent but lacks scene-dependent geometric grounding, while image-only conditioning may infer plausible motion from the scene layout but cannot reliably follow language-specified directions. Removing the dedicated <CAM> token is also expected to degrade performance, suggesting that explicit camera-aware conditioning is more effective than generic feature fusion for trajectory generation.

\subsection{Camera Trajectory Analysis via Wavelet Decomposition}
\label{appendix:wavelet_decomposition}
\begin{figure*}[t!]
    \centering
    \captionsetup{type=figure}
    \includegraphics[width=1.0\linewidth]{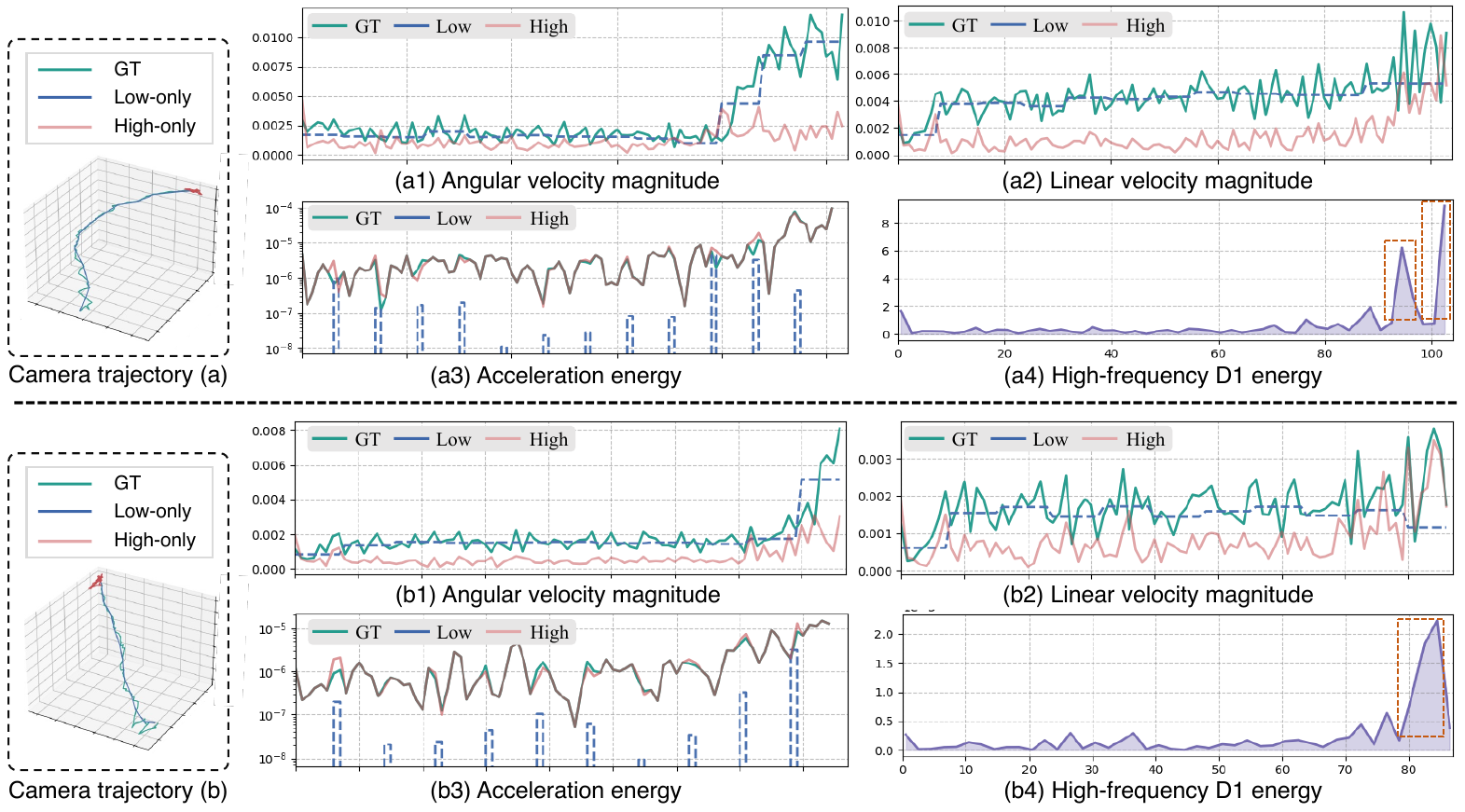}
    \caption{High- and low-frequency analysis of camera trajectories within a single scene, including angular velocity magnitude, linear velocity magnitude, acceleration energy, and high-frequency D1 energy.}
    \label{fig:wavelet_analysis_single}
\end{figure*}

We further analyze two different camera trajectories by decomposing the camera parameters into low- and high-frequency components using the wavelet transform in Fig.~\ref{fig:wavelet_analysis_single}. The following features are visualized and compared: angular velocity magnitude, linear velocity magnitude, acceleration energy, and high-frequency D1 energy. 
From these qualitative results, we can conclude that:

\begin{itemize}[leftmargin=*, noitemsep, topsep=10pt]
    \item The low-frequency component exhibits smooth variations, representing gradual rotations of the camera. In contrast, the high-frequency component displays sharp fluctuations, indicating rapid changes in angular velocity at short time scales. This behavior suggests that the high-frequency components capture sudden, localized rotations, while the low-frequency components represent overall angular motion.
    
    \item The low-frequency component captures the steady motion of the camera, reflecting its overall trajectory. The high-frequency component, on the other hand, exhibits abrupt spikes, particularly in the boxed regions, which correspond to rapid localized movements. These sharp variations highlight the high-frequency component's sensitivity to fast movements, while the low-frequency component maintains the overall motion trend.
    
    \item The low-frequency component reflects smooth and large-scale movements, while the high-frequency component emphasizes rapid and smaller movements. This decomposition illustrates how the high-frequency energy captures fast, localized accelerations, while the low-frequency component focuses on the more gradual, large-scale changes in acceleration.

    \item The high-frequency D1 energy exhibits prominent spikes, particularly in the boxed regions, indicating rapid changes in the camera's motion.
\end{itemize}

\subsubsection{Proof}
To prove the above conclusion, we describe the camera motion characteristics using kinematic models, where linear velocity \( \mathbf{v} \) and angular velocity \( \boldsymbol{\omega} \) are decomposed into their respective frequency components. The energy of the motion is computed as follows:

\begin{equation}
E = \int (\mathbf{v}^2 + \boldsymbol{\omega}^2) \, dt,
\end{equation}

\noindent where \( E \) is the total energy of the camera motion, and \( \mathbf{v} \) and \( \boldsymbol{\omega} \) are the linear and angular velocities, respectively. The energy is decomposed into low-frequency and high-frequency components through a wavelet transform, with the low-frequency components capturing the smooth, overall motion, and the high-frequency components capturing rapid, localized changes.
For the high-frequency components, we define the energy \( E_{hf} \) in the frequency domain as:

\begin{equation}
E_{hf} = \sum_{f > f_{cutoff}} |\hat{\mathbf{T}}(f)|^2,
\end{equation}

\noindent where \( \hat{\mathbf{T}}(f) \) is the Fourier transform of the camera trajectory, and \( f_{cutoff} \) is the cutoff frequency separating the low and high-frequency components. The correlation between pronounced high-frequency energy spikes and abrupt trajectory changes underscores the critical role of high-frequency components in encoding rapid temporal dynamics.

\subsection{DiT-based Camera Trajectory Modeling}
\label{appendix:dit_based_camera_modeling}
In our CT-1 model, we utilize a Diffusion Transformer to model the camera trajectories. Below, we provide further details.
Let $K_0 \in \mathbb{R}^{T \times C}$ denote a clean camera trajectory of temporal length $T$ and dimension $C$. Following the standard diffusion formulation, we define a forward noising process:

\begin{equation}
q(K_t \mid K_0) = \mathcal{N}\left(\sqrt{\bar{\alpha}_t} K_0, (1 - \bar{\alpha}_t)\mathbf{I}\right),
\end{equation}

\noindent where $K_t$ is the noisy trajectory at diffusion step $t$, and $\bar{\alpha}_t$ denotes the cumulative noise schedule. Our goal is to learn a conditional denoising model $\varepsilon_\theta(K_t, t, z)$, parameterized by a Diffusion Transformer (DiT), which predicts the noise component or an equivalent denoised representation given the camera-context token \texttt{<CAM>}.

Concretely, the noisy trajectory $K_t \in \mathbb{R}^{N \times T \times C}$ is first embedded into a latent space of dimension $D$ via a learnable projection $E_x(\cdot)$. The diffusion timestep $t$ is mapped to a global temporal embedding $E_t(t) \in \mathbb{R}^{N \times D}$, while the \texttt{<CAM>} is embedded through $E(\cdot)$ to produce $E_(cam) \in \mathbb{R}^{N \times 1 \times D}$. The timestep and \texttt{<CAM>} embeddings are fused via element-wise addition:

\begin{equation}
c = E(cam) + E(t),
\end{equation}

\noindent yielding a single \texttt{<CAM>} token that encodes both temporal and contextual information.
This \texttt{<CAM>} token is prepended to the embedded trajectory sequence, forming an augmented token sequence:

\begin{equation}
X = \big[c;E(K_t)\big] \in \mathbb{R}^{N \times (T+1) \times D},
\end{equation}

\noindent which is further enriched with learnable positional embeddings. The resulting sequence is processed by a stack of Transformer blocks, enabling global self-attention across all time steps and allowing each camera pose to directly interact with the \texttt{<CAM>} signal. Finally, a linear projection maps the Transformer outputs back to the trajectory space, producing:

\begin{equation}
\hat{K} \in \mathbb{R}^{N \times (T+1) \times C}.
\end{equation}

The first token corresponding to the \texttt{<CAM>} embedding is discarded, and the remaining $T$ tokens are used as the model output, yielding the predicted noise or denoised camera trajectory:

\begin{equation}
\varepsilon_\theta(K_t, t, z) = \hat{K}_{[:,1:,:]}.
\end{equation}

This allows the model to jointly capture long-range temporal dependencies and camera-aware trajectory dynamics, making it well-suited for camera-controllable video generation within a diffusion framework.

\subsection{Analysis of the Effect of $\beta$ in WavReg Loss.}
\label{appendix:effect_of_beta}
To train the CT-1 model, we optimize a composite objective with $\beta$:

\begin{equation}
\mathcal{L}(\theta)=\mathcal{L}_{\mathrm{diff}}(\theta)+\beta\,\mathcal{L}_{\mathrm{wav}}(\theta),
\label{eq:total_loss}
\end{equation}

\noindent where $\mathcal{L}_{\mathrm{diff}}$ is the standard diffusion training loss (\textit{e.g.,} $\epsilon$-prediction) and $\mathcal{L}_{\mathrm{wav}}$ is the proposed wavelet-domain supervision. 
The scalar $\beta>0$ controls the relative contribution of the two objectives.

\noindent \textbf{Update decomposition.}
Under stochastic gradient descent with learning rate $\eta$, the parameter update at iteration $k$ is:

\begin{equation}
\theta_{k+1}
=\theta_k-\eta\left(\nabla_\theta \mathcal{L}_{\mathrm{diff}}(\theta_k)+\beta \nabla_\theta \mathcal{L}_{\mathrm{wav}}(\theta_k)\right).
\label{eq:sgd_update}
\end{equation}

Let $g_{\mathrm{diff}}:=\nabla_\theta \mathcal{L}_{\mathrm{diff}}(\theta_k)$ and $g_{\mathrm{wav}}:=\nabla_\theta \mathcal{L}_{\mathrm{wav}}(\theta_k)$. 
Eq.~\eqref{eq:sgd_update} shows that $\beta$ directly scales the wavelet gradient component, thus changing both the \emph{magnitude} and the \emph{direction} of the update.

\noindent \textbf{Gradient-ratio (core indicator).}
To quantify the effective strength of the auxiliary wavelet supervision, we define the \emph{gradient ratio}:

\begin{equation}
r_k(\beta)
:=\frac{\left\|\beta g_{\mathrm{wav}}\right\|_2}{\left\|g_{\mathrm{diff}}\right\|_2}
=\beta \cdot \frac{\left\|g_{\mathrm{wav}}\right\|_2}{\left\|g_{\mathrm{diff}}\right\|_2}.
\label{eq:grad_ratio}
\end{equation}

In practice, due to minibatch stochasticity, we consider an expectation or moving average:

\begin{equation}
r(\beta):=\frac{\beta\,\mathbb{E}\big[\| \nabla_\theta \mathcal{L}_{\mathrm{wav}}\|_2\big]}
{\mathbb{E}\big[\|\nabla_\theta \mathcal{L}_{\mathrm{diff}}\|_2\big]}.
\label{eq:grad_ratio_expect}
\end{equation}

The ratio $r(\beta)$ characterizes the relative scale of the two gradient contributions in the update and provides a principled interpretation of $\beta$.

\noindent \textbf{Directional influence and an angle bound.}
Beyond magnitude, $\beta$ affects the \emph{direction} of the update. 
Let the combined gradient be $g_{\mathrm{tot}}:=g_{\mathrm{diff}}+\beta g_{\mathrm{wav}}$.
Consider the angle $\phi$ between $g_{\mathrm{tot}}$ and the diffusion gradient $g_{\mathrm{diff}}$:

\begin{equation}
\cos\phi
=\frac{\langle g_{\mathrm{tot}},\,g_{\mathrm{diff}}\rangle}{\|g_{\mathrm{tot}}\|_2\,\|g_{\mathrm{diff}}\|_2}
=\frac{\|g_{\mathrm{diff}}\|_2^2+\beta\langle g_{\mathrm{wav}},g_{\mathrm{diff}}\rangle}
{\|g_{\mathrm{diff}}+\beta g_{\mathrm{wav}}\|_2\,\|g_{\mathrm{diff}}\|_2}.
\label{eq:angle_def}
\end{equation}

By the triangle inequality, we have:

\begin{equation}
\|g_{\mathrm{diff}}\|_2-\|\beta g_{\mathrm{wav}}\|_2
\le
\|g_{\mathrm{tot}}\|_2
\le
\|g_{\mathrm{diff}}\|_2+\|\beta g_{\mathrm{wav}}\|_2.
\label{eq:tri_ineq}
\end{equation}

When $r_k(\beta)<1$, Eq.~\eqref{eq:tri_ineq} implies $\|g_{\mathrm{tot}}\|_2 \ge \|g_{\mathrm{diff}}\|_2-\|\beta g_{\mathrm{wav}}\|_2 = \|g_{\mathrm{diff}}\|_2(1-r_k(\beta))>0$.
Moreover, the deviation angle is bounded by:

\begin{equation}
\sin \phi
=
\frac{\|\beta g_{\mathrm{wav}} - \mathrm{proj}_{g_{\mathrm{diff}}}(\beta g_{\mathrm{wav}})\|_2}{\|g_{\mathrm{tot}}\|_2}
\le
\frac{\|\beta g_{\mathrm{wav}}\|_2}{\|g_{\mathrm{tot}}\|_2}
\le
\frac{r_k(\beta)}{1-r_k(\beta)},
\label{eq:angle_bound}
\end{equation}

\noindent where $\mathrm{proj}_{g_{\mathrm{diff}}}(\cdot)$ denotes the projection onto $g_{\mathrm{diff}}$, and the last inequality uses the lower bound in Eq.~\eqref{eq:tri_ineq}. 
Eq.~\eqref{eq:angle_bound} shows that if $r_k(\beta)\ll 1$, the update direction remains close to that induced by $\mathcal{L}_{\mathrm{diff}}$, while the wavelet term provides a controlled, bounded perturbation.

\begin{figure*}[t]
    \centering
    \captionsetup{type=figure}
    \includegraphics[width=1.0\linewidth]{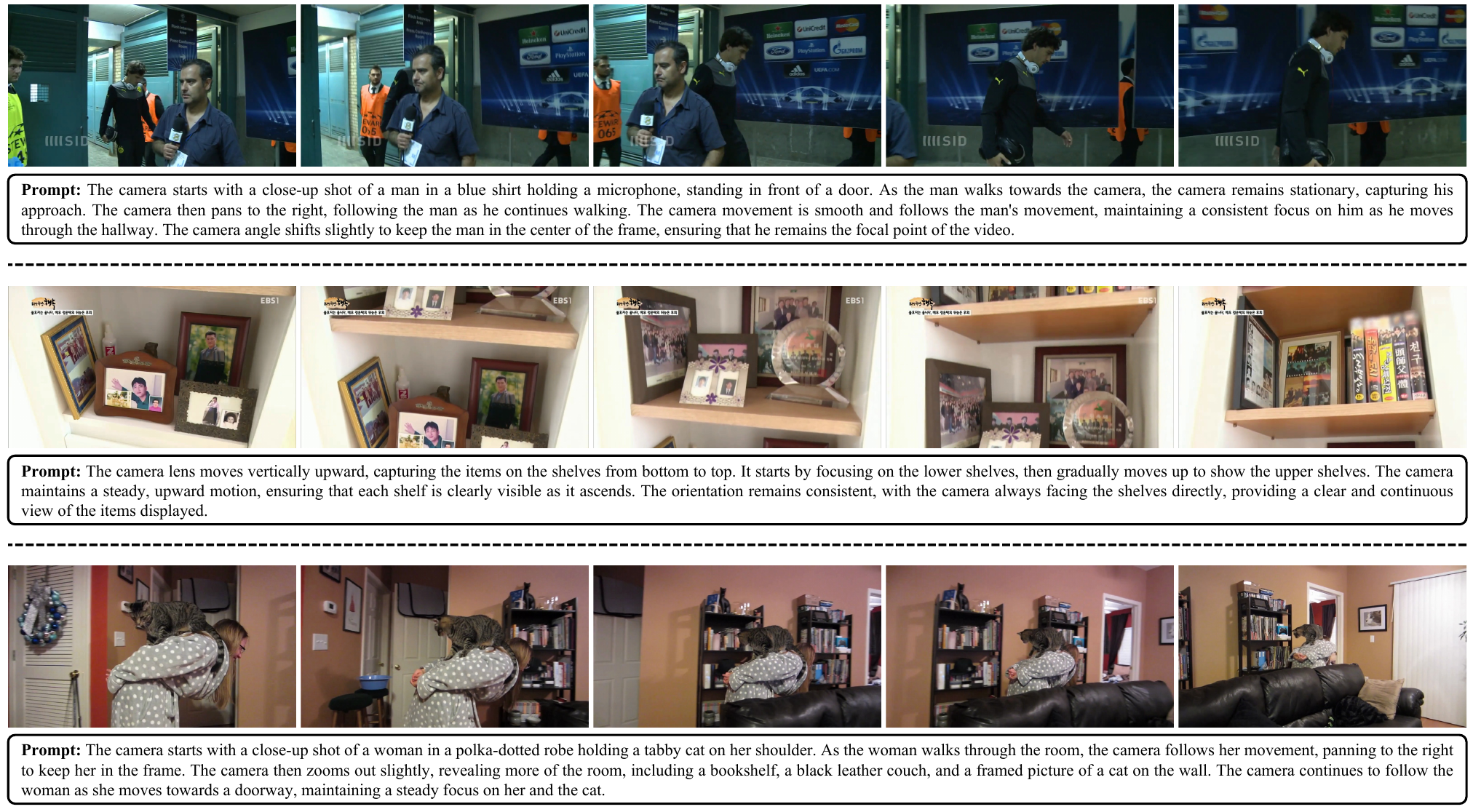}
    \caption{Example instances from the CT-200 dataset. For the raw data, we apply the data curation pipeline proposed in this work to produce fine-grained annotations of camera motion descriptions. The figure shows representative video frames along with their corresponding camera motion prompts.}
    \label{fig:dataset_show}
\end{figure*}

\subsection{CT-200K Dataset Construction}
\label{appendix:dataset}
In this section, we provide more details of our CT-200K dataset construction and we give some examples from our CT-200K in Fig.~\ref{fig:dataset_show}.

\noindent \textbf{General video scenarios.}
We first focus on generating textual annotations for general video scenarios, with the goal of creating comprehensive, well-aligned descriptions that explicitly disentangle camera motion $M_c$ from static visual content $C_s$.
Given an input video sequence $V = \{f_t\}_{t=1}^T$, where $f_t$ denotes the $t$-th frame, we first process the entire sequence using video-based vision-language models (Video-VLMs). This extracts dynamic prompts that characterize camera motion, yielding a video camera description $\mathcal{D}_v = \text{Video-VLM}(V)$. This description captures the temporal evolution of viewpoint and trajectory, $\{\Delta p_t, \Delta \theta_t\}_{t=1}^{T-1}$, over the video duration.

In parallel, individual frames are processed by image-based vision-language models (Image-VLMs) to generate static prompts describing scene content, objects, and other time-invariant visual elements. For each frame $f_t$, this forms an image content description $\mathcal{D}_i^t = \text{Image-VLM}(f_t)$.
To ensure semantic consistency, both descriptions are further refined by large language models (LLMs) guided by a fixed instruction $\mathcal{I}$:

\begin{equation}
\mathcal{D}_{\text{ref}}^t = \text{LLM}\left(\mathcal{D}_v, \mathcal{D}_i^t, \mathcal{I}\right),
\end{equation}

\noindent where $\mathcal{I}$ specifies the removal of any information in $\mathcal{D}_v$ that is not supported by $\mathcal{D}_i^t$. This filtering step eliminates hallucinated or irrelevant motion cues and enforces alignment between dynamic and static semantics, \textit{i.e.,} $\mathcal{D}_v \perp \perp \mathcal{D}_i^t \mid \mathcal{D}_{\text{ref}}^t$.
The output of this stage is a set of refined textual annotations $\{\mathcal{D}_{\text{ref}}^t\}_{t=1}^T$ for video frames, which jointly describe camera motion and visual content in a synchronized and faithful manner. This process leverages the complementary strengths of modern vision-language and language-only models, achieving performance in semantic alignment.

\noindent \textbf{Reasoning video scenarios.}
The second stage targets scenarios requiring fine-grained spatial reasoning, aiming to generate detailed textual annotations that explicitly describe object manipulations $\mathcal{A}_o$ and spatial relationships $\mathcal{R}_s$.
Given an input video sequence $V = \{f_t\}_{t=1}^T$, we first use video-based vision-language models (Video-VLMs) to extract visual cues and temporal dynamics, producing a spatio-temporal feature representation $\mathcal{F}_v = \text{Video-VLM}(V)$.
A scenario selection step then identifies video segments $\mathcal{S} = \{s_k\}_{k=1}^K$ (where $s_k = \{f_t\}_{t=t_a}^{t_b}$) that exhibit meaningful object interactions or spatial changes:

\begin{equation}
\mathcal{S} = \left\{s_k \mid \text{Scenario-Sel}\left(\mathcal{F}_v(s_k)\right) = \text{True}\right\}.
\end{equation}

For each selected segment $s_k$, large language models (LLMs) construct precise textual annotations, guided by visual evidence (including object localization cues $\mathcal{L}_o = \{\text{bbox}_o^t\}_{o,t}$ from video frames). The annotation generation process is defined as:

\begin{equation}
\mathcal{D}_{\text{spatial}}^k = \text{LLM}\left(\mathcal{F}_v(s_k), \mathcal{L}_o(s_k), \mathcal{I}_{\text{spatial}}\right).
\end{equation}

\noindent where $\mathcal{I}_{\text{spatial}}$ is a fixed instruction prompting the LLM to describe spatial transformations between objects (\textit{e.g.,} relative position changes $\{\Delta r_{o_1,o_2}^t\}$ or object movements $\{\Delta p_o^t\}$).
The resulting annotations $\{\mathcal{D}_{\text{spatial}}^k\}_{k=1}^K$ provide a structured, interpretable account of object-level interactions and spatial reasoning events within the video, delivering spatially explicit and action-oriented descriptions.

\noindent \textbf{Camera parameters estimation.}
The third stage estimates camera pose directly from video data, aiming to provide an explicit trajectory representation of camera motion $\mathcal{M}_c$ that serves as a critical component for downstream tasks requiring accurate viewpoint transformation modeling.
Given an input video sequence $V = \{f_t\}_{t=1}^T$, we feed it into a dedicated camera pose estimation model~\cite{wang2025vggt}. The model processes frame-wise geometry and inter-frame relationships to predict time-varying camera parameters.
The camera pose estimation process is defined as:

\begin{equation}
\mathcal{P}_t = \text{VGGT}(V) \quad \text{for} \quad t=1,\dots,T,
\end{equation}

\noindent where $\mathcal{P}_t = \left(I_t, R_t, \mathbf{t}_t\right)$ represents the full camera pose at frame $t$. Here, $I_t$ denotes the camera intrinsic matrix (encoding optical properties such as focal length and principal point), while $R_t$ (rotation matrix) and $\mathbf{t}_t$ (translation vector) represent the camera extrinsic parameters $K$, describing its 3D orientation and position in world coordinates.

\noindent \textbf{Statistics.} Using the data curation pipeline proposed in this work, we construct a high-quality dataset consisting of both general scenarios and reasoning scenarios. Specifically, the dataset contains \textbf{120,814} samples from General Scenarios and \textbf{80,015} samples from Reasoning Scenarios, resulting in a total of \textbf{200,829} samples. The dataset comprises over \textbf{47M} video frames, covering a wide range of scenes and dynamic motions. Each sample is paired with a corresponding textual description and an associated camera trajectory, providing rich supervision for camera-controllable video generation and trajectory modeling.